\pdfoutput=1
  
 \documentclass[11pt]{article}

  \usepackage[acceptedWithA]{tacl2021v1}

  \makeatletter
\iftaclpubformat           % only when compiled with "acceptedWithA"
\fi
\makeatother

  \usepackage{times}
 \usepackage{latexsym}
 \usepackage{xcolor}
 \usepackage{todonotes}
 
  \usepackage[T1]{fontenc}
    
  \usepackage[utf8]{inputenc}
 
    \usepackage{microtype}
 
    \usepackage{inconsolata}
 
 \usepackage{graphicx}
 \usepackage{amsmath}
 \usepackage{amssymb}
 \usepackage{amsfonts}        \usepackage{booktabs}
 \usepackage{lipsum}
 \usepackage{multicol}
 \usepackage{multirow}
 \usepackage{bm}
 \usepackage{cleveref}
 \usepackage{easyeqn}
  \usepackage{listings}
 \usepackage{placeins}
 \usepackage{color,xcolor,colortbl}
 \usepackage[multiple]{footmisc}
 \usepackage{enumitem}
 \usepackage{paralist}
 \usepackage{subcaption}
 \usepackage{xspace}
 
 \newcommand\YAMLcolonstyle{\color{black}\mdseries}
 \newcommand\YAMLkeystyle{\ttfamily\scriptsize\color{teal}\bfseries}
 \newcommand\YAMLvaluestyle{\color{black}\mdseries}
 
 \makeatletter
 
   \newcommand\language@yaml{yaml}
 
 \expandafter\expandafter\expandafter\lstdefinelanguage
 \expandafter{\language@yaml}
 {
   keywords={},
   keywordstyle=\color{black}\mdseries,
   basicstyle=\YAMLkeystyle,
   sensitive=false,
   comment=[l]{\#},
   morecomment=[s]{/*}{*/},
   commentstyle=\color{purple}\ttfamily,
   stringstyle=\YAMLvaluestyle\ttfamily,
   moredelim=[l][\color{orange}]{\&},
   moredelim=[l][\color{magenta}]{*},
   moredelim=**[il][\YAMLcolonstyle{:}\YAMLvaluestyle]{:},      morestring=[b]',
   morestring=[b]",
   literate =    {---}{{\ProcessThreeDashes}}3
                 {>}{{\textcolor{red}\textgreater}}1     
                 {|}{{\textcolor{red}\textbar}}1 
                 {\ -\ }{{\mdseries\ -\ }}3,
   belowskip=-3.0 \baselineskip
 }
 
  \lst@AddToHook{EveryLine}{\ifx\lst@language\language@yaml\YAMLkeystyle\fi}
 \makeatother
 
 \newcommand\ProcessThreeDashes{\llap{\color{cyan}\mdseries-{-}-}}

   \newcommand{\uv}{\mathbf{u}}
 \newcommand{\vv}{\mathbf{v}}
 \newcommand{\xv}{\mathbf{x}}
 \newcommand{\yv}{\mathbf{y}}
 
 \newcommand{\EE}{\mathbb{E}}
 
 \newcommand{\DC}{\mathcal{D}}
 \newcommand{\CC}{\mathcal{C}}
 \newcommand{\HC}{\mathcal{H}}

 \newcommand{\YC}{\mathcal{Y}}
 \newcommand{\Mistral}{Mistral 7B\xspace}
 \newcommand{\Stable}{Stable LM 2 12B\xspace}
 \newcommand{\PT}{P(True)\xspace}
 \newcommand{\RA}{ROC-AUC\xspace}
 \newcommand{\PA}{PR-AUC\xspace}

 \newcommand{\multirowcell}[1]{\begin{tabular}[c]{@{}c@{}}#1\end{tabular}}
 
 \newcommand\blankfootnote[1]{   \let\thefootnote\relax\footnotetext{#1}   \let\thefootnote\svthefootnote }
 
 \newcommand\blfootnote[1]{   \begingroup
   \renewcommand\thefootnote{}\footnote{#1}   \addtocounter{footnote}{-1}   \endgroup
 }

 \title{Benchmarking Uncertainty Quantification Methods\\ for Large Language Models with LM-Polygraph}

 \author{
 \bf Roman Vashurin\textsuperscript{1 $\diamondsuit$}\enspace
 Ekaterina Fadeeva\textsuperscript{2 $\diamondsuit$}\enspace
 Artem Vazhentsev\textsuperscript{3,4 $\diamondsuit$}  \\
 \bf Lyudmila Rvanova\textsuperscript{4,6}\enspace
 Daniil Vasilev\textsuperscript{5}\enspace
 Akim Tsvigun\textsuperscript{7}\enspace
 Sergey Petrakov\textsuperscript{3} \\
 \bf Rui Xing\textsuperscript{1,8}\enspace
 Abdelrahman Sadallah\textsuperscript{1}  \enspace
 Kirill Grishchenkov\textsuperscript{9}\enspace
 Alexander Panchenko\textsuperscript{3,4}\\
 \bf Timothy Baldwin\textsuperscript{1,8}\quad
 \bf Preslav Nakov\textsuperscript{1}\quad
 Maxim Panov\textsuperscript{1}\quad
 Artem Shelmanov\textsuperscript{1}\\
 \textsuperscript{1}MBZUAI \;
 \textsuperscript{2}ETH Z\"urich \;
 \textsuperscript{3}Skoltech \;
 \textsuperscript{4}AIRI \; 
 \textsuperscript{5}HSE University \;\\
 \textsuperscript{6}FRC CSC RAS \; 
 \textsuperscript{7}Nebius \; 
 \textsuperscript{8}The University of Melbourne \;
 \textsuperscript{9}Independent Researcher \;
 \\
 \href{mailto:roman.vashurin@mbzuai.ac.ae}{roman.vashurin@mbzuai.ac.ae} ~~
 \href{mailto:ekaterina.fadeeva@inf.ethz.ch}{ekaterina.fadeeva@inf.ethz.ch} ~~
 \href{mailto:vazhentsev@airi.net}{vazhentsev@airi.net} ~~ \\ \href{mailto:panchenko@airi.net}{panchenko@airi.net} \;
 \href{mailto:timothy.baldwin@mbzuai.ac.ae}{\{timothy.baldwin, preslav.nakov, maxim.panov\}@mbzuai.ac.ae} \\
 \href{mailto:artem.shelmanov@mbzuai.ac.ae}{artem.shelmanov@mbzuai.ac.ae}
 }

\begin{document}
 
 \maketitle
 
 \begin{abstract} 
      The rapid proliferation of large language models (LLMs) has stimulated researchers to seek effective and efficient approaches to deal with LLM hallucinations and low-quality outputs.
         Uncertainty quantification (UQ) is a key element of machine learning applications in dealing with such challenges. 
         However, research to date on UQ for LLMs has been fragmented in terms of techniques and evaluation methodologies.
   In this work, we address this issue by introducing a novel benchmark that implements a collection of state-of-the-art UQ baselines and offers an environment for controllable and consistent evaluation of novel UQ techniques over various text generation tasks. 
      Our benchmark also supports the assessment of confidence normalization methods in terms of their ability to provide interpretable scores.
              Using our benchmark, we conduct a large-scale empirical investigation of UQ and normalization techniques across eleven tasks, identifying the most effective approaches.
    \end{abstract}

 \section{Introduction}
 
   \blfootnote{$\diamondsuit$ Equal contribution}
 
   Uncertainty quantification (UQ) is increasingly being recognized as a critical safety component in AI applications. It enables systems to abstain from uncertain model predictions, allowing the associated inputs to be handled through alternative means -- for example, by escalating them to a human operator~\cite{el2010foundations}. This safety mechanism is crucial in areas where the cost of errors is high, such as healthcare. Besides that, uncertainty scores can be used for out-of-distribution detection (OOD)~\cite{podolskiy2021revisiting,vazhentsev2023efficient}, annotation with active learning~\cite{gal2017deep,shelmanov2021active,tsvigun2022towards,rubashevskii2023scalable}, adversarial attack detection~\cite{DBLPconfuaiSmithG18}, reducing model response latency~\cite{xin2020deebert,schwartz2020right,schuster2022confident,leviathan2023fast,chen2023accelerating} among many other applications.
    
   A plethora of UQ methods has been developed for classification and regression models~\cite{Gal2016Uncertainty}. 
   There has also been a surge of research devoted to UQ specifically in the context of encoder-only language models (LMs) such as BERT
      ~\cite{zhang-etal-2019-mitigating,he2020towards,shelmanov-etal-2021-certain,xin-etal-2021-art,vazhentsev-etal-2022-uncertainty,kotelevskii2022nonparametric,wang2022uncertainty,kuzmin-EtAl:2023:ijcnlp}. The rapid proliferation of large language models (LLMs) has stimulated researchers to seek efficient and effective approaches to UQ in text generation tasks, in an attempt to make LLMs safer to use in downstream applications.   
   As with any ML model, LLMs can make incorrect predictions, ``hallucinate'' by fabricating claims~\cite{xiao-wang-2021-hallucination,dziri-etal-2022-origin}, or simply generate low-quality outputs. These problems stem from the peculiarities of the LLM training objective, the general nature of ML models in being susceptible to errors due to the limited amount of training data, and the inherent ambiguity of tasks.

   Several methods exist for censoring the outputs of LLMs: output filtering using stop-word lists, post-processing using classifiers~\cite{xu-etal-2023-understanding}, fact-checking with external tools~\cite{wang2023factcheck}, output rewriting~\cite{logacheva-etal-2022-paradetox}, and model alignment via preference optimization~\cite{rafailov2024direct}. However, these techniques alone are insufficient to entirely eliminate incorrect/inappropriate outputs. For instance, fact-checkers target a very narrow sub-problem and usually require external knowledge sources such as knowledge bases, which are generally incomplete. Building an efficient external system to verify the LLM output for every possible task is infeasible.

   UQ offers a more general solution to the problem by relying on the model's internal capabilities without requiring access to external knowledge, which also enables the potential for greater computational efficiency.
   Several recent studies have focused on developing UQ methods for LLMs in text generation tasks~\cite{malinin2020uncertainty,van-der-poel-etal-2022-mutual,kuhn2023semantic,ren2023outofdistribution,vazhentsev2023efficient,fadeeva2023lm,lin2023generating,fadeeva2024factchecking}.
      However, the current UQ research landscape is quite fragmented, with many non-comparable and concurrent studies.
   Researchers have proposed highly divergent methods for benchmarking UQ techniques, making it challenging to consolidate research findings and draw general conclusions.
    
   In this work, we strive to bridge these disparate research efforts and resolve some issues found in their evaluation protocols by developing a benchmark for UQ techniques in text generation tasks. The benchmark is based on the LM-Polygraph framework~\cite{fadeeva2023lm}, which implements state-of-the-art UQ baselines in a unified way, enabling a large-scale, consistent comparison of methods developed in recent work. 
      It includes the tasks of selective question-answering (QA), selective generation (machine translation [MT] and text summarization [TS]), and claim-level fact-checking. For the latter, we developed an automatic fact-checking pipeline for four languages: English, Chinese, Arabic, and Russian. Besides common metrics related to UQ performance, we also introduce a metric related to the calibration of confidence scores. It enables the evaluation of confidence normalization methods according to their ability to produce interpretable scores. We propose a strong baseline for normalization and investigate its performance in comparison to simpler approaches. Using the developed benchmark, we conduct a large-scale empirical investigation of UQ and normalization methods across eleven datasets.
         
   This work both lowers the barrier to entry into UQ research for individual researchers and developers, and enables more robust, reliable, and trustworthy LLM deployment for end users.
 
   Our \textbf{contributions} are as follows:
   \begin{compactitem}
     \item We propose a new comprehensive benchmark for the evaluation of UQ and uncertainty normalization methods for LLMs. The benchmark
               can assess the calibration of uncertainty scores and their effectiveness in selective QA/generation and claim-level fact-checking (hallucination detection).\footnote{All code is published under the MIT license and available at \url{https://github.com/IINemo/lm-polygraph}
     }
          
     \item As part of the benchmark, we develop a novel multilingual automatic evaluation pipeline for claim-level UQ methods, focusing on claim-level fact-checking of LLM outputs in multiple languages, including English, Mandarin Chinese, Arabic, and Russian.

     \item We develop methods for producing normalized and bounded confidence scores that preserve the performance of raw uncertainty scores while providing better calibration and improved interpretability for end users.
 
     \item Using the developed benchmark, we perform a large-scale empirical evaluation of state-of-the-art UQ techniques.
   \end{compactitem}

 \section{Uncertainty Quantification Methods}
 
 \subsection{Background}
   Uncertainty is a fundamental concept in 
      ML
   and statistics, indicating that model predictions have a degree of variability due to the lack of complete information. Estimating predictive uncertainty is crucial for various tasks, such as selective classification, where the model abstains from making a prediction if its confidence is insufficient. 
    
   Despite recent efforts to establish a common definition of predictive uncertainty~\cite{kotelevskii2024predictive,hofman2024quantifying}, multiple approaches to its quantification exist based on probabilities, entropies, distances, risks, etc. From a practical perspective, any of these scores could serve as a measure of uncertainty as long as they accurately reflect the relevant properties and help to solve reliability tasks. 
 
   \begin{table*}[t]
\centering
\small

\scalebox{0.72}{

\begin{tabular}{clccccc}

\toprule
\textbf{Type} & \textbf{Uncertainty Quantification Method} & \textbf{Category} & \textbf{Compute} & \textbf{Memory} & \multirowcell{\textbf{Needs} \\ \textbf{Training} \\ \textbf{Data}} & \textbf{Level} \\ % Memory % Training time
\midrule

\parbox[t]{2mm}{\multirow{21}{*}{\rotatebox[origin=c]{90}{White-box}}} & Maximum Sequence Probability (MSP) & \multirow{9}{*}{\multirowcell{Information-\\based}} & Low & Low & No & Seq./claim \\
 & Perplexity \cite{fomicheva-etal-2020-unsupervised} & & Low & Low & No & Seq./claim \\
 & Mean/Max Token Entropy (TE; \citet{fomicheva-etal-2020-unsupervised}) & & Low & Low & No & Seq./claim \\ 
 & Pointwise Mutual Information (PMI; \citet{takayama-arase-2019-relevant}) & & Medium & Low & No & Seq./claim \\
 & Conditional PMI \cite{van-der-poel-etal-2022-mutual} & & Medium & Medium & No & Seq. \\
 & Rényi Divergence \cite{darrin-etal-2023-rainproof} & & Low & Low & No & Seq. \\
 & Fisher-Rao Distance \cite{darrin-etal-2023-rainproof}  & & Low & Low & No & Seq. \\
 & TokenSAR \cite{duan-etal-2024-shifting} & & Low & Low & No & Seq. \\
  & CCP \cite{fadeeva2024factchecking} & & Low & Low & No & Seq./claim \\ [00.7ex]

 & Monte Carlo Sequence Entropy (MC-SE; \citet{kuhn2023semantic}) & \multirow{5}{*}{\multirowcell{Sample \\diversity}} & High & Low & No & Seq. \\
 & Monte Carlo Norm. Seq. Entropy (MC-NSE; \citet{malinin2020uncertainty}) & & High & Low & No & Seq. \\
 & Semantic Entropy \cite{kuhn2023semantic} &  & High & Low & No & Seq. \\
 & SentenceSAR \cite{duan-etal-2024-shifting} & & High & Low & No & Seq. \\
 & SAR \cite{duan-etal-2024-shifting} & & High & Low & No & Seq. \\[00.7ex]

% Sentence-level ensemble-based measures \cite{malinin2020uncertainty} &  & \multirow{2}{*}{Ensembling} & High & High & Yes & Seq. \\
% Token-level ensemble-based measures \cite{malinin2020uncertainty} & & & High & High & Yes & Seq. \\[00.7ex]

 & Mahalanobis Distance (MD; \citet{lee2018simple}) & \multirow{4}{*}{\multirowcell{Density-\\based}} & Low & Low & Yes & Seq. \\
 & Robust Density Estimation (RDE; \citet{yoo-etal-2022-detection}) & & Low & Low & Yes & Seq. \\
 & Relative Mahalanobis Distance (RMD; \citet{ren2023outofdistribution}) & & Low & Low & Yes & Seq. \\
 & Hybrid Uncertainty Quantification (HUQ; \citet{vazhentsev-etal-2023-hybrid}) & & Low & Low & Yes & Seq. \\[00.7ex]

 & P(True) \cite{kadavath2022language} & Reflexive & Medium & Low & No & Seq./claim \\

\midrule

\parbox[t]{2mm}{\multirow{10}{*}{\rotatebox[origin=c]{90}{Black-box}}} & Number of Semantic Sets (NumSet; \citet{lin2023generating}) & \multirow{6}{*}{\multirowcell{Sample \\diversity}} & High & Low & No & Seq. \\
 & Sum of Eigenvalues of the Graph Laplacian (EigV; \citet{lin2023generating}) & & High & Low & No & Seq. \\
 & Degree Matrix (Deg; \citet{lin2023generating}) & & High & Low & No & Seq. \\
 & Eccentricity (Ecc; \citet{lin2023generating}) & & High & Low & No & Seq. \\
 & Lexical Similarity (LexSim; \citet{fomicheva-etal-2020-unsupervised}) & & High & Low & No & Seq. \\ 
 & BB Semantic Entropy & & High & Low & No & Seq.\\ [00.7ex]
 & LabelProb & Information-based & Low & Low & No & Seq.\\ [00.7ex]
 & BB P(True) & \multirow{3}{*} & Medium & Low & No & Seq./claim\\
 & Verbalized 1S \cite{tian-etal-2023-just} & {\multirowcell{Reflexive}} & Low & Low & No & Seq. \\
 & Verbalized 2S \cite{tian-etal-2023-just} & & Medium & Low & No & Seq. \\

\bottomrule

\end{tabular}

}

\caption{UQ methods implemented in the benchmark.}
\label{tab:ue_methods}

\end{table*}

         While there are principled ways of expressing and reasoning about uncertainty, e.g., in terms of information theory and Bayesian modeling~\cite{blundell2015weight}, they are often difficult to implement and may lead to worse model performance. Therefore, UQ practitioners usually rely on approximations or even heuristics.
      For example, one popular approach to UQ is ensembling~\cite{ashukha2019pitfalls}. For classification tasks, it is considered a very strong baseline, but it introduces large computational overhead due to the need for repetitive inference and storing multiple versions of weights. One of the main research questions related to UQ that has been addressed in recent work is how to perform it efficiently while keeping the performance of the uncertainty scores reliably high~\cite{shelmanov-etal-2021-certain}.
 
   UQ for text generation tasks represents a greater challenge than classification. In generation, a model makes multiple predictions: one for each token. Therefore, the uncertainty scores for each token should be somehow aggregated into a single value. At the same time, in many cases, we would like to have an uncertainty score not for the entire output but for text fragments such as individual claims.  Another problem is that the raw probability distributions of LLMs reflect multiple types of uncertainty, some of which might be irrelevant to a given generation task. Usually, we should not take into account the uncertainty related to the choice of the surface forms of the answer, as long as they convey the same meaning~\cite{kuhn2023semantic}. Similarly, uncertainty related to the type of conveyed information might be irrelevant, as long as it is correct, and we care only about its veracity~\cite{fadeeva2024factchecking}.
   Finally, LLM predictions are not conditionally independent~\cite{zhang2023enhancing}, and therefore incorrect claims generated by an LLM at the start of an output can cause flown-on hallucinations through the subsequent generation process.

 \subsection{Overview of Uncertainty Quantification Methods for LLMs}
 \label{sec:uncertainty_methods}

   Here, we provide an overview of the UQ methods implemented in our benchmark, as outlined in \Cref{tab:ue_methods}. The methods are implemented using the LM-Polygraph framework~\cite{fadeeva2023lm}, which has been extended to incorporate several recently proposed approaches. A detailed description of the methods can be found in Appendix~\ref{sec:appendix_methods}.
 
   There are two major types of techniques: white-box and black-box. \textit{White-box} methods require access to logits, internal layer outputs, or the LLM itself. \textit{Black-box} methods only need access to the generated text, and can easily be integrated with third-party online services such as OpenAI's API. Methods also differ in their computational requirements: some pose high computational or memory overheads, e.g., due to repeated inference, making them less suitable for practical usage. The application of some methods can also be hindered by the need to access the model's training data. Finally, different methods might be restricted only to the UQ of the whole text (sequence level), while others might also be applicable to text fragments, such as atomic claims (claim level).

 \subsubsection{White-Box Methods}
   Let us consider the input sequence $\xv$ and the output sequence $\yv \in \YC$ of length $L$, where $\YC$ is the set of all possible output sequences. Then the probability of an output sequence given an input sequence for autoregressive language models is given by
   \begin{equation}
     P(\yv \mid \xv) = \prod\nolimits_{l = 1}^L P(y_l \mid \yv_{<l}, \xv),
   \end{equation}
   where the distribution of each $y_l$ is conditioned on all previous tokens in a sequence $\yv_{<l} = \{y_1, \dots, y_{l - 1}\}$.
 
   We begin the discussion with \textbf{information-based methods}, which focus on analyzing token probability distributions $P(y_l \mid \yv_{<l}, \xv)$.
   The simplest UQ baseline in this group is \textit{Maximum Sequence Probability} (MSP) score:
   $U_\mathrm{MSP}(\xv) = 1 - P(\yv \mid \xv)$. MSP discards a lot of information from the LLM probability distribution, which in theory might affect the UQ performance. This issue is addressed in various \textit{entropy}-based techniques~\cite{fomicheva-etal-2020-unsupervised}. 
   \textit{Claim-Conditioned Probability} (CCP; \citet{fadeeva2024factchecking}) is another method in this category that aims to eliminate the impact of irrelevant sources of uncertainty reflected in original $P(y_l \mid \yv_{<l}, \xv)$. We will return to the discussion of CCP in detail in Section~\ref{sec:claim_level}. Information-based methods offer the advantage of being simple to implement and cost-effective, while still providing performance that is often on par with more computationally demanding techniques.
 
   Information-based methods can be improved by sampling multiple outputs from the LLM and aggregating their confidence scores or assessing their diversity. We refer to these techniques as \textbf{sample diversity} methods. 
      \citet{malinin2020uncertainty} suggest to
   sample several sequences $\yv^{(k)}, k = 1, \dots, K$ and compute entropy on the sequence level through approximate Monte Carlo estimation (\textit{Monte Carlo sequence entropy}). This approach does not take into account that many sampled responses do not diverge in meaning and only vary in surface form. This problem is addressed by \textit{Semantic Entropy} (SE; \citet{kuhn2023semantic}), which clusters sampled responses by meaning and computes entropy of the distribution over obtained clusters.
      This approach is further extended in \textit{Shifting Attention to Relevance} (SAR; \citet{duan-etal-2024-shifting}). Instead of clustering, SAR performs a soft aggregation of word or sentence probabilities using their semantic similarity. Additionally, SAR mitigates the influence of irrelevant tokens and sequence samples. 
 
   \textbf{Density-based methods}~\cite{lee2018simple,yoo-etal-2022-detection,kotelevskii2022nonparametric,ren2023outofdistribution} approximate the training data distribution using embeddings of training instances. Usually, this distribution is modeled by one or multiple Gaussians.
      Uncertainty is quantified by estimating the likelihood of the input under the approximated distribution. As such, they are good at spotting OOD instances~\cite{vazhentsev2023efficient}.  
   They are computationally efficient at inference time, with no additional inference steps required. However, they require access to the model's training data to fit the approximated distribution.
   One can also combine information-based and density-based methods as suggested by~\citet{vazhentsev-etal-2023-hybrid} and~\citet{ren2023outofdistribution}. For example, the \textit{Hybrid Uncertainty Quantification} (HUQ) method~\cite{vazhentsev-etal-2023-hybrid} performs a ranking-based aggregation and leverages the strengths of both information-based and density-based methods.
    
   Directly asking the model to provide confidence for its responses is another approach to UQ~\cite{kadavath2022language,tian-etal-2023-just}. We refer to such techniques as \textbf{reflexive}. In one of the simplest techniques of this kind \PT~\citep{kadavath2022language}, the authors generate a response from an LLM and subsequently ask the same LLM to verify the answer. The uncertainty score is calculated using the probability of the option ``True'' in the output distribution of the LLM for the second prompt. \citet{kadavath2022language} showed that this method achieves better performance than MSP at the cost of an additional inference step on a variety of tasks for large LLMs ($\geq$ 70b parameters).

 \subsubsection{Black-Box Methods}
   LLM providers often expose models in a black-box fashion, where users have access to generated text only.    Among \textbf{sample diversity} black-box methods (also known as consistency-based methods in the black-box setting~\cite{zhang2024luq}), we consider Lexical Similarity~\cite{fomicheva-etal-2020-unsupervised}, Number of Semantic Sets, Sum of Eigenvalues of the Graph Laplacian, Degree Matrix, and Eccentricity. 
      These methods are grounded in a common methodological framework~\cite{lin2023generating}:
   \begin{compactitem}
     \item Obtain $K$ responses $\yv^{(1)}, \dots, \yv^{(K)}$ for a particular input $\xv$.
     \item Compute $K \times K$ similarity matrix $S$ between responses, where $S_{ij} = s\bigl(\yv^{(i)}, \yv^{(j)}\bigr)$ for some similarity score $s$ (e.g., NLI or Jaccard score).
     \item Analyze the similarity matrix $S$ and aggregate the information in this matrix to compute the resulting uncertainty score.
   \end{compactitem}
         \textit{Number of Semantic Sets} is the simplest method, which clusters semantically similar responses into non-overlapping groups and counts the resulting clusters. A larger number of clusters indicates greater uncertainty. Other methods do a more sophisticated analysis of the matrix $S$. For example, \textit{Sum of Eigenvalues of the Graph Laplacian} computes the sum of eigenvalues of the normalized matrix $S$, providing a continuous relaxation of the Number of Semantic Sets score.
 
   \textbf{Reflexive} techniques for the black-box setting are 
      enabled by the fact that most popular LLMs deployed as a service are instruction-tuned and are
      able to follow a multi-turn conversation.
      This allows the user to either prompt a model to explicitly verbalize its confidence level as part of its response or request a confidence estimate in a follow-up conversation turn. \citet{tian-etal-2023-just} propose several variations of such \textit{Verbalized} UQ methods and conduct an extensive empirical evaluation.

 \subsubsection{Claim-Level Extensions}
 \label{sec:claim_level}
   While the methods discussed above provide uncertainty scores for entire generated sequences, it is often desirable to quantify the uncertainty for short text fragments (claims) within the LLM output. 
      Assuming that claims have been extracted from text sentences and there is a mapping between them and the tokens in the original text, we can obtain probability distributions for each token in each claim. Some aforementioned sequence-level methods can be modified to operate on the claim level~\citep{fadeeva2024factchecking}, but not all of them. For example, sampling-based methods cannot work on the claim level because sampled texts may diverge too much and miss some claims.
 
   Let $C$ denote a set of token indices corresponding to a claim. To adapt \textit{MSP} to the claim level, we can compute the joint probability of tokens solely within the claim instead of the whole sequence: $\prod_{l \in C} P(y_l \mid \yv_{<l})$. In a similar way, we can adapt \textit{Mean/Max Token Entropy}, \textit{Perplexity}, and \textit{PMI}. 
      \textit{\PT} could be adapted to the claim level by querying an LLM about the correctness of each claim in the generated response individually.
    
      \textit{Claim-Conditioned Probability} (CCP; \citet{fadeeva2024factchecking}) is designed specifically for the claim-level UQ (but can also be applied at the sequence level). It assesses the semantic similarity between the original claim and perturbed versions where individual tokens are replaced with alternative generations. This approach provides a more nuanced understanding of uncertainty by considering the potential impact of different word choices on the overall meaning of the claim.

 \section{Uncertainty Normalization Methods}
   \label{sec:normalization_theory_short}
   Raw uncertainty scores are good for ranking outputs by their potential quality but can be confusing to the end user. To address this issue, we consider several methods for producing confidence scores bounded within the range $[0,1]$.
 
   All methods require fitting on a held-out calibration set $\mathcal{D}_{calib} = \{\xv_i, \yv_i^*\}_{i=1}^N$. We assume that for each input $\xv_i$ in this set, output $\yv_i$ is generated by LLM, and some uncertainty score $u_i = U(\xv_i)$ as well as the generation quality score $q_i = Q(\yv_i, \yv_i^*)$ are computed.
 
   Among simple normalization approaches, we consider \textit{linear scaling} and \textit{quantile scaling}, as they provide simple rules to normalize uncertainty scores in $[0, 1]$ based solely on uncertainty values $u_i$; see Appendix~\ref{sec:normalization_theory} for more details.
 
   To convey meaningful information about the model's confidence to the end user, the confidence score should not only be bounded within a fixed interval but also directly reflect the expected quality of the model's output. We term this \textit{confidence calibration}, as opposed to confidence normalization. To achieve this, we introduce two methods referred to as \textit{Performance-Calibrated Confidence} (PCC). 
 
   The first approach, \textit{Binned PCC}, splits the calibration set into non-intersecting bins based on the values of uncertainty $u_i$ and considers the confidence to be an estimate of the mean quality of the outputs in the bin, as measured by some quality measure of choice. The downside of this approach is that the ordering of the instances based on raw uncertainty and normalized confidence scores can be different, and thus the quality of UQ can vary substantially and unpredictably.
   
   To address this problem, we propose a second approach: \textit{Isotonic PCC}. It fits Centered Isotonic Regression (CIR; \citet{Oron2017CenteredIR}) on pairs of uncertainty and quality values from the calibration set. CIR produces a monotonic piecewise linear function, which allows the use of the relationship between uncertainty and quality while keeping the order of the inputs intact.
   
   Both approaches in the PCC family produce calibrated confidence scores as a local estimate of some quality measure in the neighborhood of the raw uncertainty estimate. This directly ties the confidence with the estimated quality of the output, thus making it more interpretable than raw uncertainty scores.
            We provide a more detailed discussion on the specifics of these methods in Appendix~\ref{sec:normalization_theory}.

 \section{Approaches to Evaluating Uncertainty Quantification Methods}
   In general, a valid UQ technique should produce scores that are well correlated with some measure of output quality. Thus, the most straightforward way of comparing different UQ methods is to measure the rank correlation between some generation quality metric (e.g., ROUGE-L) and uncertainty scores~\cite{fomicheva-etal-2020-unsupervised, ren2023outofdistribution}. However, this way of evaluating UQ is not very informative of the performance gain that a particular UQ method achieves. 
 
   Another popular evaluation approach is based on designating outputs as either correct or incorrect based on a threshold over a quality metric, and measuring how well uncertainty scores can predict the output as being one or the other~\cite{kuhn2023semantic,duan-etal-2024-shifting}. This reduces the uncertainty score to being a predictor in a binary classification task, and thus one can use \RA or \PA as a measure of how well a UQ method performs. The problem with this approach, which makes results across different works incomparable, is that it requires selecting the quality threshold, and its choice has been quite arbitrary in the literature.
 
   A more comprehensive approach is called \textit{rejection verification}~\cite{malinin2020uncertainty, lin2023generating}. It does not require thresholding the quality metric to formulate a binary classification task. Instead, it computes the average quality of the outputs for which the uncertainty is relatively low. By continuously lowering the uncertainty threshold above which data points are discarded, one obtains a set of average quality values of outputs with progressively lower maximum uncertainty. These pairs of uncertainty thresholds and associated average output quality give a prediction--rejection curve. The area under this curve quantifies the overall quality of an UQ method~\cite{malinin-etal-2017-incorporating}. 
 
   One problem with the majority of evaluation approaches in previous work on UQ for LLMs is the usage of $n$-gram output quality metrics such as ROUGE-L, which often do not reflect the actual quality of the generated output. For example, this discrepancy occurs when the gold standard answer and the LLM output differ only by a negation token. In this case, $n$-gram metrics would rank such a model answer higher than it deserves, failing to capture the substantial semantic difference caused by the negation. In addition to $n$-gram metrics, we suggest using the recently proposed AlignScore~\cite{zha-etal-2023-alignscore}, where the gold-standard answer and LLM output are compared by another LLM. This metric has a higher correlation with human annotators due to its ability to capture deeper semantic similarities and differences between texts. 
 
   Another issue with evaluation protocols in recent works is their tendency to overlook simple yet effective baselines. For instance, MSP often proves difficult to surpass in tasks with short outputs. Many studies neglect this baseline, favoring comparisons with entropy-based metrics~\cite{he2020towards,xiao-wang-2021-hallucination,kuhn2023semantic}, which often perform worse. By not considering this straightforward baseline, the evaluation protocols may give an incomplete picture of UQ performance.
 
   We also note that research on UQ for text fragments, such as sentences and claims, has been limited. Furthermore, the evaluation protocols for this setting exhibit significant limitations. 
      For example, \citet{manakul2023selfcheckgpt} manually annotated texts generated by one LLM, then evaluated the UQ performance for a proxy model by inferring the probability distributions of the tokens for the fixed output. 
      We argue that such an approach introduces a big discrepancy between the generated text and what a proxy LLM actually wants to generate, which results in biased UQ performance. In this work, we mitigate this problem by building an automatic evaluation pipeline for UQ in claim-level fact-checking for various languages. It allows unrestricted generation from LLMs and leverages LLM-as-a-judge~\cite{zheng2023judging} instead of manual annotation to support experiments with various LLMs.

   \begin{table}[t]
\centering
\resizebox{0.48\textwidth}{!}{\begin{tabular}{l|cccccc}
\toprule
\textbf{Dataset} & \textbf{Type} & \multirowcell{\textbf{Num. Instances} \\ \textbf{train/test}} & \multirowcell{\textbf{Avg.} \\ \textbf{Document} \\ \textbf{Len.}} & \multirowcell{\textbf{Avg.} \\ \textbf{Target} \\ \textbf{Len.}} & \textbf{Language} &  \\
\hline

%\multicolumn{5}{c}{\textbf{QA}} \\
% CoQA & QA & 7199 / 500 / - & 405.9 & 4 & English \\ 
% TriviaQA & QA & 138384 / 17944 / 17210 & 18.8 & 4.3 & English \\ 
% MMLU & QA & 99842 / 1531 / 14042 & 64.9 & 3 & English \\ 
% GSM8k & QA & 7473 / - / 1319 & 64.2 & 128.6 & English \\ 
% XSum & ATS & 204045 / 11332 / \textbf{11334} & 544.2 & 30.4 & English \\
CoQA & QA & 7,199 / 500  & 405.9 & 4 & English \\ 
TriviaQA & QA & 138,384 / 17,210 & 18.8 & 4.3 & English \\ 
MMLU & QA & 99,842 / 14,042 & 64.9 & 3 & English \\ 
GSM8k & QA & 7,473 / 1,319 & 64.2 & 128.6 & English \\ 
XSum & ATS & 204,045 / 11,334 & 544.2 & 30.4 & English \\
%bAbiQA & 2000 / - / \textbf{200} & 31.1 & 1.0 & English \\ 
%\multicolumn{5}{c}{\textbf{NMT}} \\
% WMT'14 & NMT & 40.8M / 3000 / \textbf{3003} & 49.3 & 32.9 & Fr.-to-Eng. \\
% WMT'19 & NMT & 34.8M / \textbf{2998} / -  & 52.5 & 33.5 & Ger.-to-Eng. \\
WMT'14 & NMT & 40.8M / 3,003 & 49.3 & 32.9 & Fr.-to-Eng. \\
WMT'19 & NMT & 34.8M / 2,998 & 52.5 & 33.5 & Ger.-to-Eng. \\
%\multicolumn{5}{c}{\textbf{ATS}} \\
%AESLC & 14436 / 1960 / \textbf{1906} & 165.5 & 6.7 & English \\ 
\bottomrule
\end{tabular}}
%\caption{}
\caption{The statistics of the benchmark datasets. The lengths in tokens are provided according to the Mistral 7B v0.2 tokenizer.}

\label{tab:datasets_stats}
\end{table}

 \section{Evaluation Benchmark}
 \label{sec:benchmark}
   Our benchmark features three sections: (1) evaluation of UQ performance in selective QA/generation; (2) evaluation of UQ performance in fact-checking; (3) evaluation of confidence calibration.

 \subsection{Selective QA / Generation}
   In this section of the benchmark, we evaluate how well the uncertainty scores of the considered methods detect low-quality LLM generations.
 
 \noindent\textbf{Datasets.} 
   For selective QA, we use four datasets: CoQA~\cite{coqa} with free-form answers about conversations; TriviaQA~\cite{joshi-etal-2017-triviaqa} with complex and compositional questions without context; MMLU~\cite{mmlu} -- a multitask dataset structured as multiple-choice QA; and GSM8k~\cite{gsm8k}, consisting of grade school math word problems. The first two datasets have been used widely in UQ research, while the last two are popular datasets for evaluating the English-language generation quality of LLMs. For selective generation, we use two MT datasets: WMT-14 French to English~\cite{wmt14} and WMT-19 German to English~\cite{barrault-etal-2019-findings}, and one text summarization dataset: XSum~\cite{xsum}. 
   
   For each dataset, we limit the evaluation set to 2,000 instances, except for MMLU, where we restrict the number of questions to 100 per subject. We also reserve 1,000 instances from the training set for UQ techniques that require ``pre-training'', such as density-based methods. The detailed statistics about the datasets are given in Table~\ref{tab:datasets_stats}.
   
   We primarily use prompt formats from the lm-evaluation-harness framework~\cite{eval-harness}. For TriviaQA, MMLU, and GSM8k, we use a 5-shot prompt. For XSum, WMT-14 Fr-En, and WMT-19 De-En, we use a zero-shot prompt. For CoQA, we use a few-shot prompt with all preceding questions in the conversation before the target question. The maximum length of the generated sequence in selective QA and generation tasks was set to the 99th percentile of the target sequence length in the respective training set. 
 
   The datasets of the benchmark could be found in the Huggingface repository\footnote{\url{https://huggingface.co/LM-Polygraph}}.

 \noindent\textbf{Metrics.}
   Following previous work on UQ in text generation~\cite{malinin2020uncertainty,vazhentsev-etal-2022-uncertainty}, we compare the methods using the Prediction Rejection Ratio (PRR) metric~\cite{malinin-etal-2017-incorporating}. Consider a test dataset $\DC = \{(\xv_j, \yv_j^*)\}$. Let $\yv_j$ be the output generated by an LLM for an input $\xv_j$ and $u_j = U(\xv_j)$ be the uncertainty score of a prediction. 
   The rejection curve indicates how the average quality $Q(\yv_j, \yv_j^*)$ of the instances with uncertainty $u_j < a$ depends on the value of the rejection parameter $a$. 
   PRR computes the ratio of the area  between the rejection curves for a considered uncertainty score and a random score and the area between the oracle (the best possible uncertainty that sorts instances according to their text quality metric) and a random score:
   \begin{equation}
     \label{eq:prr}
     PRR = \frac{\text{AUC}_{\text{unc}}-\text{AUC}_{\text{rnd}}}{\text{AUC}_{\text{oracle}}-\text{AUC}_{\text{rnd}}}.
   \end{equation}
      A higher PRR indicates a better uncertainty score.
   
   The choice of the generation quality measure $Q(\yv_j, \yv^*_j)$ depends on the dataset. In contrast to previous work that employs $n$-gram-based measures, our benchmark primarily relies on LLM-based metrics, such as AlignScore~\cite{zha-etal-2023-alignscore} and COMET~\cite{rei-etal-2020-comet}, while retaining $n$-gram measures for comparability with prior research.
   For MT datasets, we use COMET and AlignScore. For GSM8K and MMLU, we use accuracy. For CoQA and TriviaQA, we use AlignScore. For XSum, we use ROUGE-L and AlignScore. 
 
   When calculating AlignScore for QA outputs, we treat the model's generations as context and a ground truth response as the claim. This approach is designed to accurately score instances where the model mentions the correct entity in its response but includes an additional explanation or rephrases the correct answer differently from the ground truth, which would otherwise lead to failure in exact match scoring. This is especially important in evaluation on the CoQA dataset.
    
   \begin{table}[!t]
\resizebox{0.49 \textwidth}{!}{
    \centering
    \footnotesize
    
    \begin{tabular}{l|cccc|c}
    \toprule
    \textbf{Language} & \textbf{Acc.} & \multirowcell{\textbf{F1} \\ \textbf{score}} & \textbf{\# claims} & \multirowcell{\textbf{\% of} \\ \textbf{false} \\ \textbf{claims}} & \multirowcell{\textbf{Fleiss} \\ \textbf{Kappa}}\\
    \midrule
    English, Mistral-7B-v0.1& 0.98 & 0.93 & 97 & 16.5\% & 0.90 \\
    Arabic, Jais 13B & 0.89 & 0.80 & 132 & 28.3\% & 0.86 \\
    Russian, Vikhr 7B-0.2 & 0.89 & 0.80 & 275 & 15.6\% & 0.85 \\
    Chinese, Yi 7B& 0.89 & 0.89 & 100 & 35.0\% & 0.87 \\
    \bottomrule
    \end{tabular}
}
    \caption{\label{tab:claim_level_human_anno} Classification metrics of GPT-4 annotation against manual annotation in claim-level fact-checking  (unsupported claims represent a positive class) and annotation agreement (Fleiss Kappa) using True/False labels from 3 annotators. }
\end{table}

% English texts are generated using Mistral-7B-v0.1, Arabic texts -- using Jais 13B, Chinese texts -- using Yi 7B~\cite{Yi2023}, Russian texts -- using Vikhr 7B-0.2~\cite{nikolich2024vikhr}. 

 \subsection{Claim-Level Fact-Checking}
   In this part of the benchmark, we evaluate claim-level UQ techniques and their ability to spot hallucinations on the task of generating biographies, as proposed by~\citet{fadeeva2024factchecking}. The difference over previous work such as~\cite{manakul2023selfcheckgpt} is that in our benchmark, we perform unrestricted generation using the LLM, which is much closer to the practical use case. However, this raises the problem that each generation is unique and needs to be reannotated, posing a significant challenge for human annotation.
 
 \noindent\textbf{Evaluation pipeline.}
   We implement an automatic benchmarking pipeline to enable the evaluation of UQ methods on unrestricted LLM outputs. The pipeline supports English, Mandarin Chinese, Arabic, and Russian. It is intended to work with any modern LLM that has functionality over these languages. The main feature of our pipeline is its full automation without needing human labeling at any step. The automation is achieved via extensive use of GPT-4. Note that GPT-4 is used only as a tool for benchmarking, while UQ is performed solely based on the particular LLM in question.
 
   Using a LLM, e.g. Mistral 7b, we first generate responses to biography prompts such as \textit{Give me a biography of <person name>}. The set of people was generated by asking GPT-4 to list the most famous people since 1900. The maximum output length is 256 tokens. 
      The generated texts are then decomposed into atomic claims, with each claim mapped to a corresponding subset of tokens from the original LLM output.
      The decomposition and mapping are done using GPT-4 with a language-specific prompt. Usually, about 5\% of claims cannot be mapped to tokens because GPT-4 abstains from responding or outputs words not present in the original text. The further evaluation considers only successfully matched claims.
 
   The annotation of the extracted claims is also done automatically using GPT-4. We use language-specific prompts that facilitate chain-of-thought reasoning to ask whether the presented claims are supported, unsupported, or unknown. Usually, the percentage of claims classified as unknown is negligible; these claims are discarded from the evaluation.
   To assess the quality of automatic annotation, we manually annotated a random subset of claims for each language.\footnote{Human annotations are available at \\\url{https://huggingface.co/datasets/LM-Polygraph/bio-claim-human-anno}}  Table~\ref{tab:claim_level_human_anno} summarizes the binary performance metrics of GPT-4 against human labels and presents annotation agreement scores. The results indicate that GPT-4 is a reliable evaluator, capable of serving as a ``ground truth'' for assessing UQ techniques.

 \noindent\textbf{Metrics.}
   The performance of the claim-level UQ methods is evaluated using \RA and \PA with unsupported claims as a positive class.

   \begin{figure*}[t!]
     \centering
     \includegraphics[trim={0.cm 0.cm 0.cm 0.cm},clip,width=0.90\linewidth]{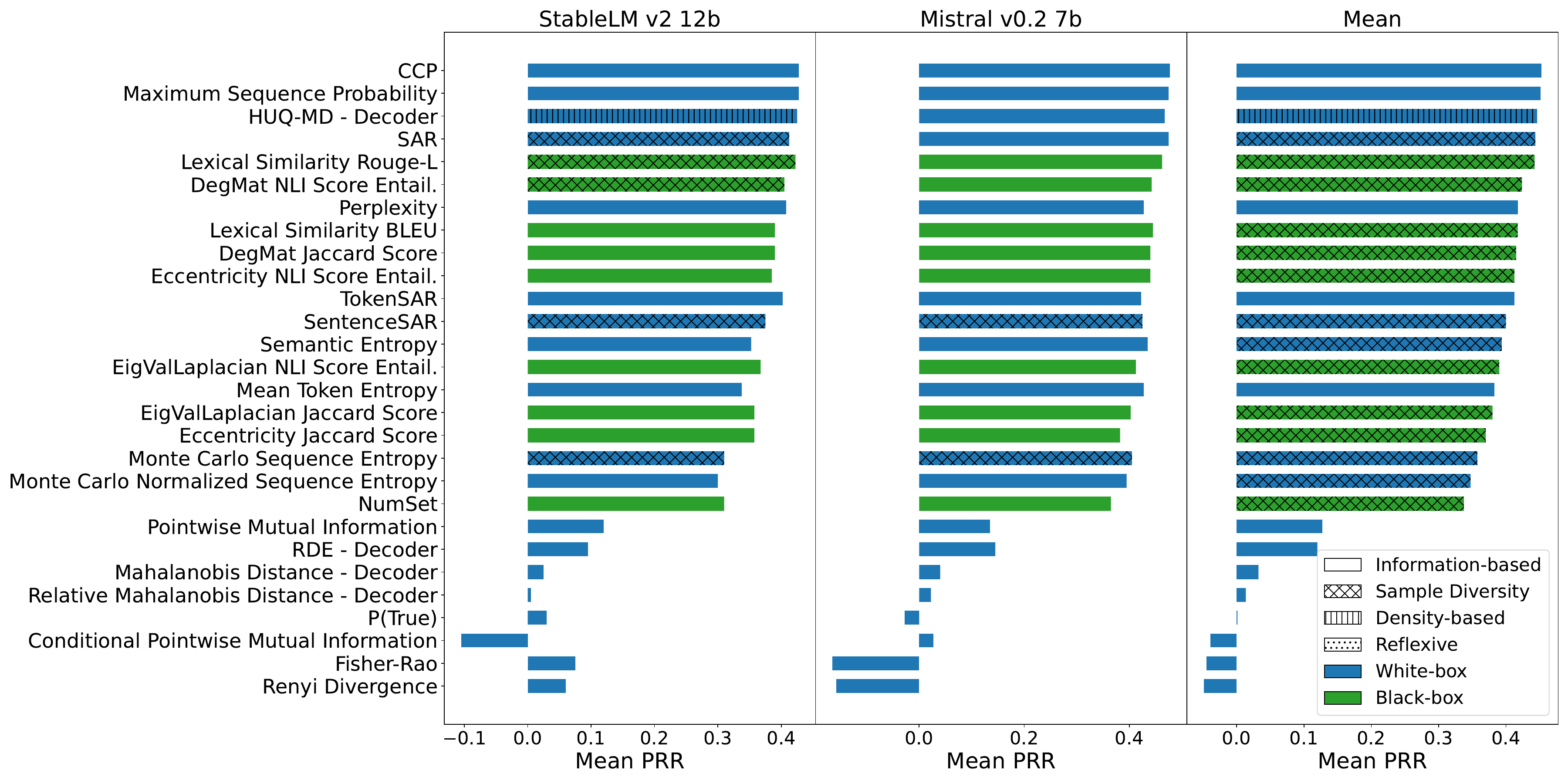}
     \caption{
       Mean PRR $\uparrow$ aggregated over all selective QA tasks for each white-box LLM separately (the lower the better). Column \emph{Mean} corresponds to the mean PRR  across all LLMs.
     }
     \label{fig:selective_classification_ranks}
   \end{figure*}
 
   \begin{figure*}[t!]
     \centering
     \includegraphics[trim={0.cm 0.cm 0.cm 0.cm},clip,width=0.90\linewidth]{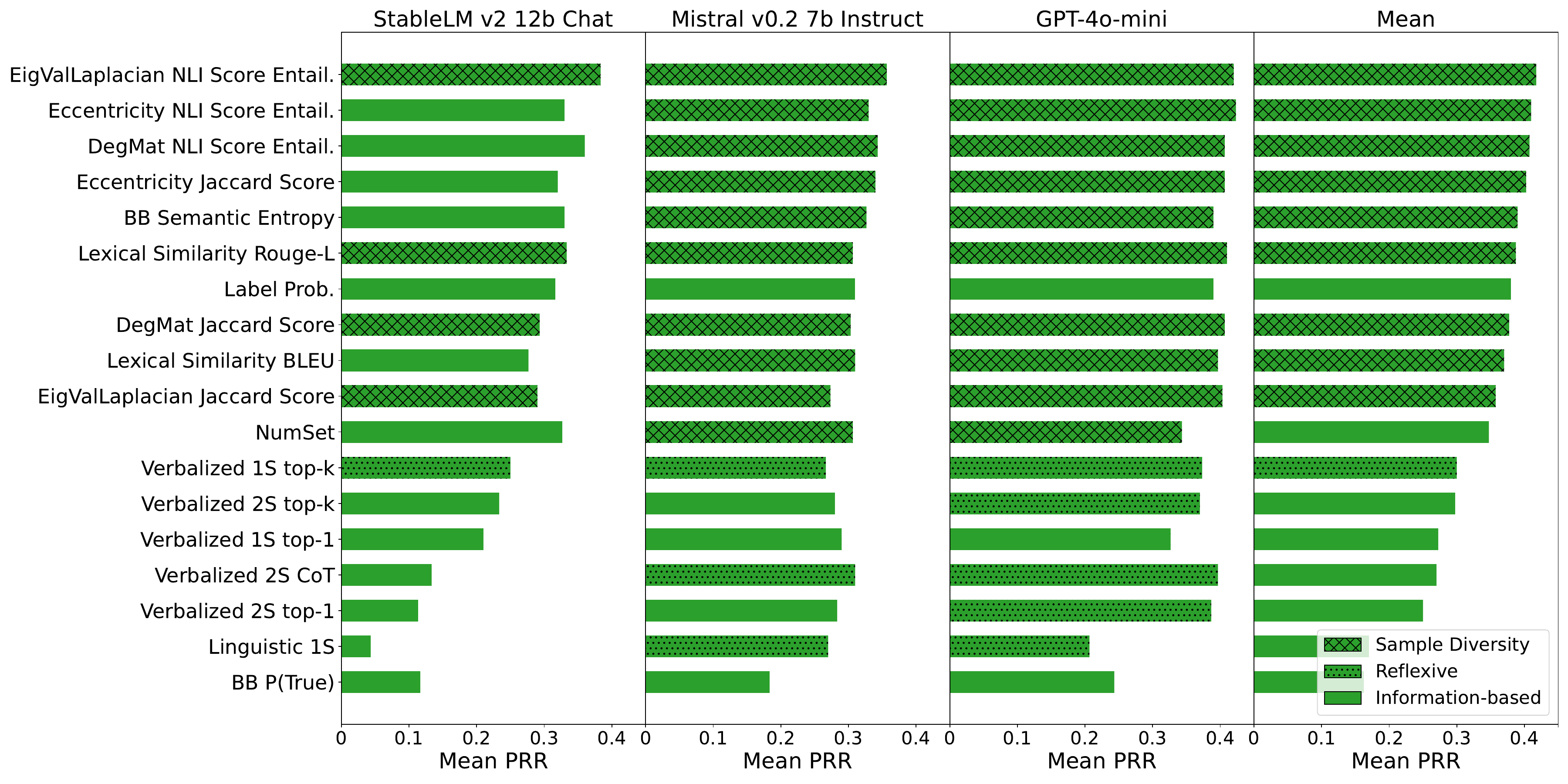}
     \caption{
       Mean PRR $\uparrow$ aggregated over all selective QA tasks for each black-box LLM separately (the lower the better). Column \emph{Mean} corresponds to the mean PRR across all LLMs.
     }
     \label{fig:blackbox_selective_classification_ranks}
   \end{figure*}
 
   \begin{figure*}[t!]
     \centering
     \includegraphics[trim={0.cm 0.cm 0.cm 0.cm},clip,width=0.90\linewidth]{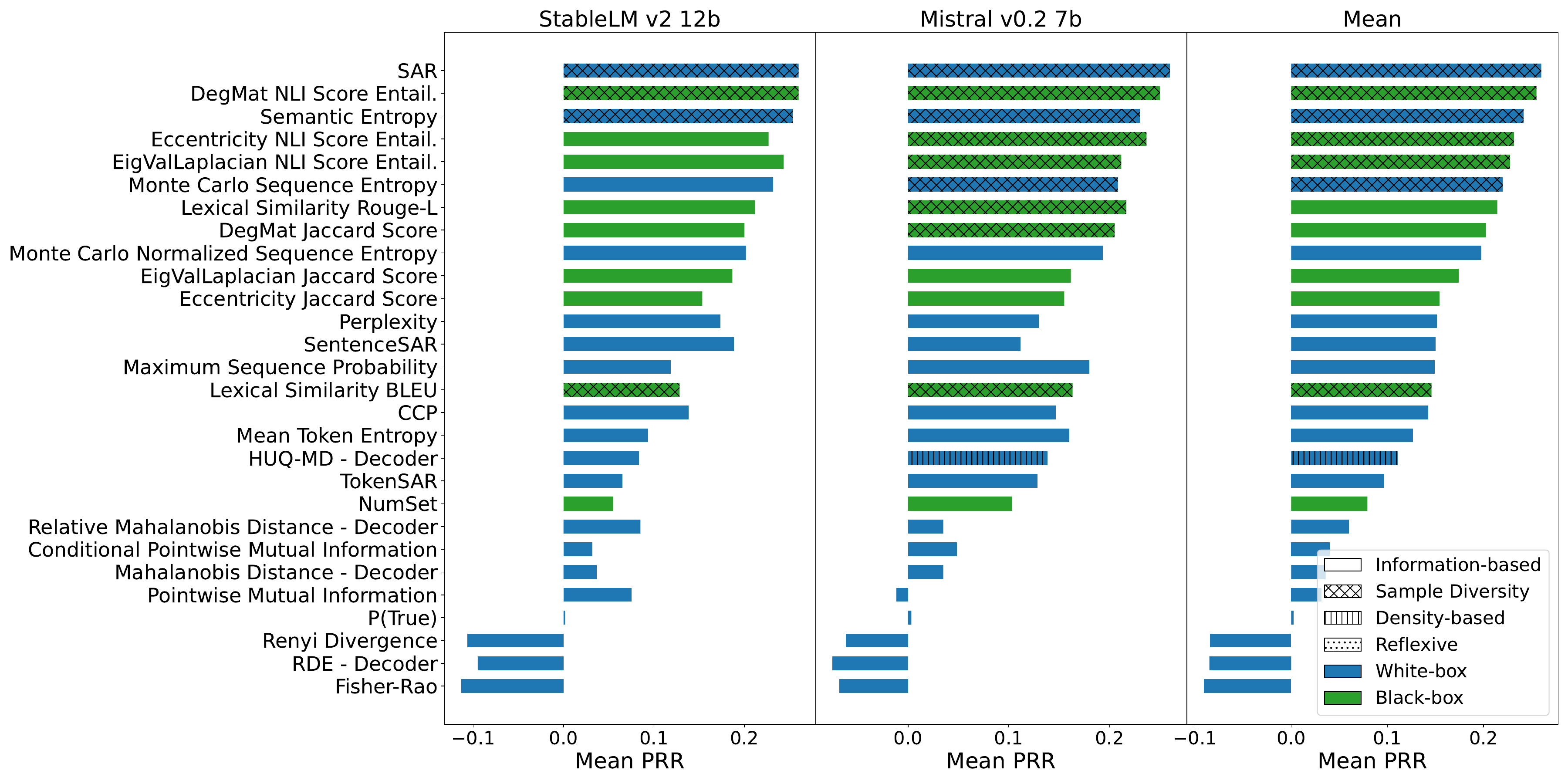}
     \caption{
       The mean PRR $\uparrow$ aggregated over all selective generation tasks for each LLM separately (the lower the better). Column \emph{Mean} corresponds to the mean PRR across all LLMs.
     }
     \label{fig:selective_generation_ranks}
   \end{figure*}

 \subsection{Effect of Uncertainty Normalization}
   This section of the benchmark is designed to analyze how uncertainty normalization procedures (as presented in Section~\ref{sec:normalization_theory_short}) affect the performance of the scores and their correlation with the quality of the generated text.
 
 \noindent\textbf{Datasets.}
   The reserved training partitions of the datasets from the selective QA and generation section are used as calibration sets for normalization methods.
         Evaluation can be performed either on the concatenated test partitions or on each of them individually.
 
 \noindent\textbf{Metrics.}
   The benchmark offers two metrics: PRR before and after normalization to ensure that performance does not degrade; and a metric similar to ECE that measures ``calibration'', i.e.\ the ability of normalized uncertainty scores to represent the expected quality of the output in a bounded range -- Mean Squared Error (MSE) between a normalized quality metric and a confidence score.    Lower MSE indicates better confidence ``calibration''.

 \subsection{Models} 
   For selective QA / generation, we conducted experiments with white-box models that provide access to logits and their internal states. For selective QA, we also conducted experiments with black-box models that provide only the generated text. Black-box models represent the scenario where LLMs are deployed as services and are available only via an API (e.g. ChatGPT and models deployed on platforms like HuggingFace).
   For selective QA/generation in the white-box setting, we use \Mistral v0.2~\cite{mistral} and \Stable~\cite{bellagente2024stable} base models without instruction tuning. For black-box evaluation on selective QA, we use the corresponding instruction-tuned versions of these models and also GPT-4o-mini. The experiments on claim-level fact-checking are conducted using the instruction-tuned versions of \Mistral-v0.1~\cite{mistral} (for English), Jais 13B~\cite{sengupta2023jais} (for English and Arabic), Vikhr 7B-0.2~\cite{nikolich2024vikhr} (for Russian), and Yi 7B (for Chinese). 
         The detailed generation hyperparameters can be found in the code base.
   The text generation quality of the models is presented in Table~\ref{tab:stable_mistral} in Appendix~\ref{sec:quality_metrics}.
 
      For white-box models, we use continuation-style prompts. For black-box models with verbalized UQ techniques, we use prompts specified by~\citet{tian-etal-2023-just}, and for other black-box UQ methods, we use general instruction-oriented prompts.
      For both model types, we compute the same metrics. However, for black-box models with verbalized UQ methods, we perform more extensive output post-processing to extract the model's predictions and disentangle them from reported confidence..

 \section{Experiments}
 \label{sec:results}
   Using our benchmark, we evaluated the implemented UQ and normalization methods.

 \subsection{Selective QA and Generation}
   \noindent\textbf{Selective QA.} Tables~\ref{tab:stablelm_cls_results} and~\ref{tab:mistral_cls_results} in Appendix~\ref{sec:additional_exps} present detailed results on the selective QA task for white-box models. \Cref{fig:selective_classification_ranks,fig:blackbox_selective_classification_ranks} present the aggregated results.
 
   Despite being a simplistic baseline, MSP demonstrates strong performance. On MMLU, it is the best method for Stable LM and the second best for Mistral. It achieves the second-best result on CoQA for Mistral and outperforms entropy-based methods in most cases.
      For GSM8k, MSP substantially lags behind the best techniques, but still outperforms entropy-based methods for Mistral. 
    
         Among information-based methods, it is also worth noting CCP as it demonstrates the best performance on MMLU and GSM8k for Mistral and has the second best result on TriviaQA and MMLU for Stable LM.
 
   The majority of density-based methods demonstrate poor results across all tasks.  One exception is HUQ-MD, which is a hybrid method that leverages strengths from both information- and density-based approaches. HUQ-MD delivers strong performance across all tasks and LLMs, achieving the best results on GSM8k for Stable LM with a substantial improvement over the nearest competitor.
 
   State-of-the-art methods based on sample diversity typically perform well on CoQA and TriviaQA. The black-box method based on the degree matrix -- DegMat with NLI similarity metric, achieves the best results for both LLMs. However, for MMLU, 
      these methods fall significantly behind information-based techniques. 
   This can be attributed to the nature of the multiple-choice QA task in MMLU, where LLMs are constrained to choose from a limited set of options and generate very short responses, limiting their capacity to exploit sample diversity.
   For GSM8k, the results are mixed: methods based on sample diversity perform similarly to information-based approaches. In the sample diversity group, it is worth also highlighting SAR, which demonstrates strong overall performance, achieving the second-best result on GSM8k and the best result on TriviaQA for Mistral. 
 
   The reflexive method \PT in most cases is not better than random. This could be due to the used LLMs being too small to develop an awareness of their own knowledge gaps, a capability observed in larger LLMs~\cite{kadavath2022language}.

   Figure~\ref{fig:selective_classification_ranks} presents the mean PRRs for each white-box model individually, as well as the mean PRR across both models. The best methods overall in selective QA for white-box models are CCP, MSP, HUQ-MD, SAR, and Lexical Similarity.

   \Cref{tab:gpt4o_cls_results,tab:stablelm_instruct_cls_results,tab:mistral_instruct_cls_results} in Appendix~\ref{sec:additional_exps} present evaluation results for the instruction-tuned LLMs treated as black-box models. 
      For all LLMs, the pattern is similar. On the CoQA and TriviaQA datasets, empirical information-based and sample diversity methods confidently outperform reflexive techniques: \PT and verbalized uncertainty. 
      On MMLU, this pattern is reversed: verbalized UQ methods notably outperform other techniques. 
      For GPT-4o-mini, improvements are consistent across the majority of verbalized uncertainty methods. For Stable LM and Mistral, they are still the best, though many of them perform comparably to sample diversity techniques. 
       
   Averaging results across all datasets and models (see Figure~\ref{fig:blackbox_selective_classification_ranks}) still places sample diversity-based methods at the top.
      Three best techniques in this experiment -- EigValLaplacian, DegMat, and Eccentricity use the NLI-based similarity measure for capturing the semantic diversity of responses.
    
   \begin{table}[t] \footnotesize
\centering\resizebox{0.48\textwidth}{!}{\begin{tabular}{l|r|r|r|r|r|r|r}
\toprule
\textbf{UQ Method} & \multicolumn{7}{c}{\textbf{Output length in symbols}} \\\cline{2-8}
 & 1--2 & 3--6 & 7--24 & 25--85 & 86--138 & 139--210 & 211--1k \\\midrule

MSP & \cellcolor[rgb]{0.852836579,0.50777808,0.575116406} 0.71 & \cellcolor[rgb]{0.8616576191882352,0.5344491213176471,0.5814942527764706} 0.62 & \cellcolor[rgb]{0.8898725387078432,0.6051525891921569,0.6035518578607844} 0.67 & \cellcolor[rgb]{0.9441952453705882,0.708851458745098,0.6639489555019608} 0.57 & \cellcolor[rgb]{0.9659156483,0.7595427616,0.7032398043} 0.54 & \cellcolor[rgb]{0.9691631781666666,0.9044582760156863,0.8705807575137254} 0.38 & \cellcolor[rgb]{0.9418435698882353,0.9280538589764706,0.9201288350882354} 0.26 \\
Perplexity & \cellcolor[rgb]{0.852836579,0.50777808,0.575116406} 0.71 & \cellcolor[rgb]{0.8871684250764706,0.5998796340235294,0.6012672772176471} 0.59 & \cellcolor[rgb]{0.8675383126470588,0.5522298155294117,0.585746150627451} 0.71 & \cellcolor[rgb]{0.9497671903627452,0.7203459010784313,0.6720534316176471} 0.56 & \cellcolor[rgb]{0.9849255762058824,0.8479150297647058,0.7906558870392157} 0.41 & \cellcolor[rgb]{0.9176723556764705,0.9302569986470588,0.9494852049705882} 0.27 & \cellcolor[rgb]{0.9418435698882353,0.9280538589764706,0.9201288350882354} 0.26 \\
Mean Token Entropy & \cellcolor[rgb]{0.8790560841823529,0.5840607685176471,0.5944135352882353} 0.68 & \cellcolor[rgb]{0.9763803588352942,0.8914823988,0.8493228856529411} 0.30 & \cellcolor[rgb]{0.947942297417647,0.7165372783529411,0.6693403172588235} 0.54 & \cellcolor[rgb]{0.9791396989627451,0.8021675484411765,0.7416485506941177} 0.47 & \cellcolor[rgb]{0.9841016995,0.8604220499999999,0.8061464956666666} 0.39 & \cellcolor[rgb]{0.8863529743019607,0.9194891086196078,0.9746593799568628} 0.22 & \cellcolor[rgb]{0.8792694128431373,0.9163932970294117,0.9792039273627451} 0.17 \\
% MC$-$SE & \cellcolor[rgb]{0.9261890675039215,0.6732459732470588,0.6401732343490196} 0.62 & \cellcolor[rgb]{0.9834610525,0.8259843225764706,0.7659985578117647} 0.39 & \cellcolor[rgb]{0.9385746672999999,0.6973227293333333,0.6558619128666667} 0.57 & \cellcolor[rgb]{0.9575785619705882,0.7384632635882353,0.686089707382353} 0.54 & \cellcolor[rgb]{0.9077541933039216,0.6388904935882354,0.6201467828333334} 0.68 & \cellcolor[rgb]{0.9607031106137255,0.7457102085921569,0.6917042176882353} 0.60 & \cellcolor[rgb]{0.9754778064901961,0.8934375166666666,0.8523803414019608} 0.33 \\
% MC$-$NSE & \cellcolor[rgb]{0.9261890675039215,0.6732459732470588,0.6401732343490196} 0.62 & \cellcolor[rgb]{0.9696268857588235,0.769790744282353,0.7119501024647059} 0.45 & \cellcolor[rgb]{0.8898725387078432,0.6051525891921569,0.6035518578607844} 0.67 & \cellcolor[rgb]{0.9575785619705882,0.7384632635882353,0.686089707382353} 0.54 & \cellcolor[rgb]{0.9560162876490197,0.7348397910862745,0.6832824522294118} 0.57 & \cellcolor[rgb]{0.9819030282176471,0.8170942072647058,0.7568604245764706} 0.52 & \cellcolor[rgb]{0.9824176791176471,0.8723068372941176,0.8216194376764706} 0.36 \\
PMI & \cellcolor[rgb]{0.9841016995,0.8604220499999999,0.8061464956666666} 0.44 & \cellcolor[rgb]{0.9236824528058823,0.9312362411882353,0.942770235182353} 0.19 & \cellcolor[rgb]{0.9377786937156862,0.9301210794313726,0.9257150330490196} 0.11 & \cellcolor[rgb]{0.61490285,0.649358983,0.8768415764999999} $-$0.24 & \cellcolor[rgb]{0.661859207627451,0.7165792202745098,0.9323611776862746} $-$0.28 & \cellcolor[rgb]{0.6194742294509804,0.6561827473294117,0.8828379510882353} $-$0.17 & \cellcolor[rgb]{0.7338390473411764,0.8027956158117646,0.9842731405470588} 0.0 \\
Conditional PMI & \cellcolor[rgb]{0.9024823794117647,0.9258330802784314,0.9630825372156863} 0.26 & \cellcolor[rgb]{0.815044265017647,0.8762581198529411,0.9992540061705882} 0.04 & \cellcolor[rgb]{0.7825907906117646,0.8497192224705882,0.9983175350588236} $-$0.21 & \cellcolor[rgb]{0.6918310328862745,0.7550917086431372,0.9589153366156863} $-$0.12 & \cellcolor[rgb]{0.7554121621254901,0.8246983074117646,0.9925393881882354} $-$0.12 & \cellcolor[rgb]{0.6569731755647059,0.7100258308470588,0.927496270517647} $-$0.11 & \cellcolor[rgb]{0.6355521478235294,0.6800053309882352,0.9035475637176471} $-$0.12 \\
Rényi Divergence & \cellcolor[rgb]{0.7311771982901961,0.7999150558117647,0.9829285958960784} 0.05 & \cellcolor[rgb]{0.8230564053823529,0.8822182482647059,0.9984342312529412} 0.05 & \cellcolor[rgb]{0.61490285,0.649358983,0.8768415764999999} $-$0.54 & \cellcolor[rgb]{0.9024823794117647,0.9258330802784314,0.9630825372156863} 0.18 & \cellcolor[rgb]{0.9377786937156862,0.9301210794313726,0.9257150330490196} 0.21 & \cellcolor[rgb]{0.6918310328862745,0.7550917086431372,0.9589153366156863} $-$0.06 & \cellcolor[rgb]{0.61490285,0.649358983,0.8768415764999999} $-$0.15 \\
Fisher-Rao Distance & \cellcolor[rgb]{0.7554121621254901,0.8246983074117646,0.9925393881882354} 0.08 & \cellcolor[rgb]{0.8309839479705883,0.8877457334411765,0.9969682626882352} 0.06 & \cellcolor[rgb]{0.7285232392627451,0.797002774964706,0.9815146148450979} $-$0.31 & \cellcolor[rgb]{0.7258692802352941,0.794090494117647,0.9801006337941176} $-$0.07 & \cellcolor[rgb]{0.8792694128431373,0.9163932970294117,0.9792039273627451} 0.09 & \cellcolor[rgb]{0.8816813900509803,0.917546110909804,0.9778288382784314} 0.21 & \cellcolor[rgb]{0.9236824528058823,0.9312362411882353,0.942770235182353} 0.23 \\
TokenSAR & \cellcolor[rgb]{0.8616576191882352,0.5344491213176471,0.5814942527764706} 0.70 & \cellcolor[rgb]{0.8871684250764706,0.5998796340235294,0.6012672772176471} 0.59 & \cellcolor[rgb]{0.8734190061058824,0.5700105097411765,0.5899980484784314} 0.70 & \cellcolor[rgb]{0.9497671903627452,0.7203459010784313,0.6720534316176471} 0.56 & \cellcolor[rgb]{0.9844312501823529,0.8554192419058824,0.7999502522156863} 0.40 & \cellcolor[rgb]{0.9112102024705883,0.9284487582705883,0.9553975652941177} 0.26 & \cellcolor[rgb]{0.9418435698882353,0.9280538589764706,0.9201288350882354} 0.26 \\
CCP & \cellcolor[rgb]{0.9441952453705882,0.708851458745098,0.6639489555019608} 0.59 & \cellcolor[rgb]{0.852836579,0.50777808,0.575116406} 0.63 & \cellcolor[rgb]{0.9150932609745098,0.6523663817764705,0.6274457140333334} 0.62 & \cellcolor[rgb]{0.9575785619705882,0.7384632635882353,0.686089707382353} 0.54 & \cellcolor[rgb]{0.9607031106137255,0.7457102085921569,0.6917042176882353} 0.56 & \cellcolor[rgb]{0.9838554631666667,0.8314865062313725,0.7721615929176471} 0.50 & \cellcolor[rgb]{0.9824176791176471,0.8723068372941176,0.8216194376764706} 0.36 \\
MC$-$SE & \cellcolor[rgb]{0.9261890675039215,0.6732459732470588,0.6401732343490196} 0.62 & \cellcolor[rgb]{0.9834610525,0.8259843225764706,0.7659985578117647} 0.39 & \cellcolor[rgb]{0.9385746672999999,0.6973227293333333,0.6558619128666667} 0.57 & \cellcolor[rgb]{0.9575785619705882,0.7384632635882353,0.686089707382353} 0.54 & \cellcolor[rgb]{0.9077541933039216,0.6388904935882354,0.6201467828333334} 0.68 & \cellcolor[rgb]{0.9607031106137255,0.7457102085921569,0.6917042176882353} 0.60 & \cellcolor[rgb]{0.9754778064901961,0.8934375166666666,0.8523803414019608} 0.33 \\
MC$-$NSE & \cellcolor[rgb]{0.9261890675039215,0.6732459732470588,0.6401732343490196} 0.62 & \cellcolor[rgb]{0.9696268857588235,0.769790744282353,0.7119501024647059} 0.45 & \cellcolor[rgb]{0.8898725387078432,0.6051525891921569,0.6035518578607844} 0.67 & \cellcolor[rgb]{0.9575785619705882,0.7384632635882353,0.686089707382353} 0.54 & \cellcolor[rgb]{0.9560162876490197,0.7348397910862745,0.6832824522294118} 0.57 & \cellcolor[rgb]{0.9819030282176471,0.8170942072647058,0.7568604245764706} 0.52 & \cellcolor[rgb]{0.9824176791176471,0.8723068372941176,0.8216194376764706} 0.36 \\
Semantic Entropy & \cellcolor[rgb]{0.9126469050843138,0.6478744190470588,0.6250127369666667} 0.64 & \cellcolor[rgb]{0.9696268857588235,0.769790744282353,0.7119501024647059} 0.45 & \cellcolor[rgb]{0.8952807659705883,0.6156984995294118,0.6081210191470588} 0.66 & \cellcolor[rgb]{0.9102005491941176,0.643382456317647,0.6225797599} 0.63 & \cellcolor[rgb]{0.8844643114450981,0.5946066788549021,0.5989826965745099} 0.72 & \cellcolor[rgb]{0.9218513348607843,0.6650340647294117,0.635032405109804} 0.68 & \cellcolor[rgb]{0.9791396989627451,0.8021675484411765,0.7416485506941177} 0.44 \\
SentenceSAR & \cellcolor[rgb]{0.8790560841823529,0.5840607685176471,0.5944135352882353} 0.68 & \cellcolor[rgb]{0.9028614815235294,0.6299065681294118,0.6152808287} 0.57 & \cellcolor[rgb]{0.8675383126470588,0.5522298155294117,0.585746150627451} 0.71 & \cellcolor[rgb]{0.9622044108411765,0.7492947789176471,0.6945295061352941} 0.53 & \cellcolor[rgb]{0.9640580048333334,0.9110985744313725,0.882570674627451} 0.28 & \cellcolor[rgb]{0.8863529743019607,0.9194891086196078,0.9746593799568628} 0.22 & \cellcolor[rgb]{0.9236824528058823,0.9312362411882353,0.942770235182353} 0.23 \\
SAR & \cellcolor[rgb]{0.8871684250764706,0.5998796340235294,0.6012672772176471} 0.67 & \cellcolor[rgb]{0.9175136022176471,0.6568221562117647,0.6298915758705882} 0.55 & \cellcolor[rgb]{0.8557769257294118,0.5166684271058823,0.5772423549254901} 0.73 & \cellcolor[rgb]{0.9028614815235294,0.6299065681294118,0.6152808287} 0.64 & \cellcolor[rgb]{0.9077541933039216,0.6388904935882354,0.6201467828333334} 0.68 & \cellcolor[rgb]{0.9283579338254901,0.6773519275058824,0.6427436489686275} 0.67 & \cellcolor[rgb]{0.9460687713941176,0.7126943685490196,0.6666446363803922} 0.52 \\
MD $-$ Decoder & \cellcolor[rgb]{0.7907430740941177,0.8567252977647059,0.9991571764705882} 0.12 & \cellcolor[rgb]{0.6944259359764706,0.7581492177882353,0.9606867415411764} $-$0.11 & \cellcolor[rgb]{0.6426363887647059,0.690064711317647,0.9117342756235294} $-$0.48 & \cellcolor[rgb]{0.6766845796941177,0.736034329145098,0.9462852021294117} $-$0.14 & \cellcolor[rgb]{0.6892991246352941,0.7519281085960785,0.9568458054235294} $-$0.23 & \cellcolor[rgb]{0.61490285,0.649358983,0.8768415764999999} $-$0.18 & \cellcolor[rgb]{0.6217599191764706,0.6595946294941176,0.885836138382353} $-$0.14 \\
RDE $-$ Decoder & \cellcolor[rgb]{0.6426363887647059,0.690064711317647,0.9117342756235294} $-$0.06 & \cellcolor[rgb]{0.8910245585529412,0.9214321063294117,0.9714899216352941} 0.14 & \cellcolor[rgb]{0.9563825307352941,0.9183409471764705,0.8972560558823529} 0.16 & \cellcolor[rgb]{0.6996157421568627,0.7642642360784314,0.9642295513921568} $-$0.11 & \cellcolor[rgb]{0.6944259359764706,0.7581492177882353,0.9606867415411764} $-$0.22 & \cellcolor[rgb]{0.7690021078509803,0.8374507968235294,0.9958609467764705} 0.05 & \cellcolor[rgb]{0.8123516188058824,0.8741592436176471,0.9993597327901961} 0.09 \\
RMD $-$ Decoder & \cellcolor[rgb]{0.61490285,0.649358983,0.8768415764999999} $-$0.10 & \cellcolor[rgb]{0.7988883877470588,0.8636648624411765,0.9998883658882353} 0.02 & \cellcolor[rgb]{0.866948893627451,0.9100089393823529,0.9853621830490196} $-$0.05 & \cellcolor[rgb]{0.876805309,0.9151164254999999,0.9804355785000001} 0.14 & \cellcolor[rgb]{0.8956961428039216,0.9233751040392157,0.9683204633137255} 0.12 & \cellcolor[rgb]{0.7125994850980393,0.779529089882353,0.9730307285392157} $-$0.03 & \cellcolor[rgb]{0.6217599191764706,0.6595946294941176,0.885836138382353} $-$0.14 \\
HUQ-MD $-$ Decoder & \cellcolor[rgb]{0.9196824685392158,0.6609281104705882,0.632461990490196} 0.63 & \cellcolor[rgb]{0.9423217193470588,0.7050085489411765,0.6612532746235295} 0.51 & \cellcolor[rgb]{0.9807976965156863,0.8111235437352942,0.7507756750235295} 0.40 & \cellcolor[rgb]{0.7934605019215686,0.8590606561960784,0.9994370569411765} 0.02 & \cellcolor[rgb]{0.8096589725941177,0.8720603673823529,0.9994654594098039} $-$0.03 & \cellcolor[rgb]{0.7392311256470588,0.8082818218039216,0.9863604477156862} 0.01 & \cellcolor[rgb]{0.8230564053823529,0.8822182482647059,0.9984342312529412} 0.10 \\
P(True) & \cellcolor[rgb]{0.6497206297058824,0.7001240916470588,0.9199209875294118} $-$0.05 & \cellcolor[rgb]{0.61490285,0.649358983,0.8768415764999999} $-$0.22 & \cellcolor[rgb]{0.6286169883529411,0.6698302759882353,0.8948307002647058} $-$0.51 & \cellcolor[rgb]{0.6194742294509804,0.6561827473294117,0.8828379510882353} $-$0.23 & \cellcolor[rgb]{0.61490285,0.649358983,0.8768415764999999} $-$0.37 & \cellcolor[rgb]{0.626331298627451,0.6664183938235294,0.8918325129705882} $-$0.16 & \cellcolor[rgb]{0.707400451427451,0.7734367635137256,0.9695437661686275} $-$0.03 \\
NumSet & \cellcolor[rgb]{0.9385746672999999,0.6973227293333333,0.6558619128666667} 0.60 & \cellcolor[rgb]{0.9528917390058824,0.727592846082353,0.6776679419235294} 0.49 & \cellcolor[rgb]{0.9150932609745098,0.6523663817764705,0.6274457140333334} 0.62 & \cellcolor[rgb]{0.9849255762058824,0.8479150297647058,0.7906558870392157} 0.41 & \cellcolor[rgb]{0.9796923648137255,0.8051528802058824,0.7446909254705882} 0.48 & \cellcolor[rgb]{0.9845960255235294,0.8529178378588236,0.7968521304901961} 0.47 & \cellcolor[rgb]{0.8792694128431373,0.9163932970294117,0.9792039273627451} 0.17 \\
EigValLaplacian NLI Entail. & \cellcolor[rgb]{0.9196824685392158,0.6609281104705882,0.632461990490196} 0.63 & \cellcolor[rgb]{0.9240202011823531,0.6691400189882354,0.6376028197294119} 0.54 & \cellcolor[rgb]{0.8675383126470588,0.5522298155294117,0.585746150627451} 0.71 & \cellcolor[rgb]{0.852836579,0.50777808,0.575116406} 0.71 & \cellcolor[rgb]{0.852836579,0.50777808,0.575116406} 0.77 & \cellcolor[rgb]{0.8587172724588235,0.5255587742117647,0.5793683038509804} 0.77 & \cellcolor[rgb]{0.8704786593764706,0.5611201626352941,0.5878720995529412} 0.62 \\
EigValLaplacian NLI Contra. & \cellcolor[rgb]{0.9646785691470589,0.756126767372549,0.7003363715784314} 0.55 & \cellcolor[rgb]{0.9367011412764705,0.6934798195294117,0.6531662319882353} 0.52 & \cellcolor[rgb]{0.9028614815235294,0.6299065681294118,0.6152808287} 0.65 & \cellcolor[rgb]{0.9404481933235294,0.7011656391372549,0.658557593745098} 0.58 & \cellcolor[rgb]{0.9528917390058824,0.727592846082353,0.6776679419235294} 0.58 & \cellcolor[rgb]{0.9844470791666666,0.8397397817137255,0.7814061455764706} 0.49 & \cellcolor[rgb]{0.915574114,0.9297565972666666,0.9515550793333334} 0.22 \\
EigValLaplacian Jaccard & \cellcolor[rgb]{0.9385746672999999,0.6973227293333333,0.6558619128666667} 0.60 & \cellcolor[rgb]{0.9528917390058824,0.727592846082353,0.6776679419235294} 0.49 & \cellcolor[rgb]{0.8844643114450981,0.5946066788549021,0.5989826965745099} 0.68 & \cellcolor[rgb]{0.9218513348607843,0.6650340647294117,0.635032405109804} 0.61 & \cellcolor[rgb]{0.9004151256333333,0.6254146054,0.6128478516333333} 0.69 & \cellcolor[rgb]{0.9326956664686274,0.6855638360235294,0.6478844782078431} 0.66 & \cellcolor[rgb]{0.9528917390058824,0.727592846082353,0.6776679419235294} 0.51 \\
DegMat NLI Entail. & \cellcolor[rgb]{0.9126469050843138,0.6478744190470588,0.6250127369666667} 0.64 & \cellcolor[rgb]{0.9102005491941176,0.643382456317647,0.6225797599} 0.56 & \cellcolor[rgb]{0.852836579,0.50777808,0.575116406} 0.74 & \cellcolor[rgb]{0.852836579,0.50777808,0.575116406} 0.71 & \cellcolor[rgb]{0.852836579,0.50777808,0.575116406} 0.77 & \cellcolor[rgb]{0.852836579,0.50777808,0.575116406} 0.78 & \cellcolor[rgb]{0.8616576191882352,0.5344491213176471,0.5814942527764706} 0.63 \\
DegMat NLI Contra. & \cellcolor[rgb]{0.9834610525,0.8259843225764706,0.7659985578117647} 0.48 & \cellcolor[rgb]{0.9240202011823531,0.6691400189882354,0.6376028197294119} 0.54 & \cellcolor[rgb]{0.9150932609745098,0.6523663817764705,0.6274457140333334} 0.62 & \cellcolor[rgb]{0.9404481933235294,0.7011656391372549,0.658557593745098} 0.58 & \cellcolor[rgb]{0.9528917390058824,0.727592846082353,0.6776679419235294} 0.58 & \cellcolor[rgb]{0.9845960255235294,0.8529178378588236,0.7968521304901961} 0.47 & \cellcolor[rgb]{0.9024823794117647,0.9258330802784314,0.9630825372156863} 0.20 \\
DegMat Jaccard & \cellcolor[rgb]{0.9126469050843138,0.6478744190470588,0.6250127369666667} 0.64 & \cellcolor[rgb]{0.9305268001470588,0.6814578817647059,0.6453140635882353} 0.53 & \cellcolor[rgb]{0.8734190061058824,0.5700105097411765,0.5899980484784314} 0.7 & \cellcolor[rgb]{0.9175136022176471,0.6568221562117647,0.6298915758705882} 0.62 & \cellcolor[rgb]{0.8952807659705883,0.6156984995294118,0.6081210191470588} 0.7 & \cellcolor[rgb]{0.9218513348607843,0.6650340647294117,0.635032405109804} 0.68 & \cellcolor[rgb]{0.9283579338254901,0.6773519275058824,0.6427436489686275} 0.55 \\
Eccentricity NLI Entail. & \cellcolor[rgb]{0.9385746672999999,0.6973227293333333,0.6558619128666667} 0.60 & \cellcolor[rgb]{0.947942297417647,0.7165372783529411,0.6693403172588235} 0.50 & \cellcolor[rgb]{0.8675383126470588,0.5522298155294117,0.585746150627451} 0.71 & \cellcolor[rgb]{0.8587172724588235,0.5255587742117647,0.5793683038509804} 0.7 & \cellcolor[rgb]{0.852836579,0.50777808,0.575116406} 0.77 & \cellcolor[rgb]{0.8587172724588235,0.5255587742117647,0.5793683038509804} 0.77 & \cellcolor[rgb]{0.852836579,0.50777808,0.575116406} 0.64 \\
Eccentricity NLI Contra. & \cellcolor[rgb]{0.9646785691470589,0.756126767372549,0.7003363715784314} 0.55 & \cellcolor[rgb]{0.9367011412764705,0.6934798195294117,0.6531662319882353} 0.52 & \cellcolor[rgb]{0.8898725387078432,0.6051525891921569,0.6035518578607844} 0.67 & \cellcolor[rgb]{0.9404481933235294,0.7011656391372549,0.658557593745098} 0.58 & \cellcolor[rgb]{0.9560162876490197,0.7348397910862745,0.6832824522294118} 0.57 & \cellcolor[rgb]{0.9845960255235294,0.8529178378588236,0.7968521304901961} 0.47 & \cellcolor[rgb]{0.9090282467058823,0.927794838772549,0.9573188082745099} 0.21 \\
Eccentricity Jaccard & \cellcolor[rgb]{0.9385746672999999,0.6973227293333333,0.6558619128666667} 0.60 & \cellcolor[rgb]{0.9659156483,0.7595427616,0.7032398043} 0.46 & \cellcolor[rgb]{0.8952807659705883,0.6156984995294118,0.6081210191470588} 0.66 & \cellcolor[rgb]{0.9348276152529411,0.6896369097254902,0.6504705511098039} 0.59 & \cellcolor[rgb]{0.9077541933039216,0.6388904935882354,0.6201467828333334} 0.68 & \cellcolor[rgb]{0.9441952453705882,0.708851458745098,0.6639489555019608} 0.64 & \cellcolor[rgb]{0.9622044108411765,0.7492947789176471,0.6945295061352941} 0.49 \\
Lexical Similarity Rouge-L & \cellcolor[rgb]{0.9126469050843138,0.6478744190470588,0.6250127369666667} 0.64 & \cellcolor[rgb]{0.9175136022176471,0.6568221562117647,0.6298915758705882} 0.55 & \cellcolor[rgb]{0.8675383126470588,0.5522298155294117,0.585746150627451} 0.71 & \cellcolor[rgb]{0.9028614815235294,0.6299065681294118,0.6152808287} 0.64 & \cellcolor[rgb]{0.8844643114450981,0.5946066788549021,0.5989826965745099} 0.72 & \cellcolor[rgb]{0.9028614815235294,0.6299065681294118,0.6152808287} 0.71 & \cellcolor[rgb]{0.9126469050843138,0.6478744190470588,0.6250127369666667} 0.57 \\
Lexical Similarity BLEU & \cellcolor[rgb]{0.9126469050843138,0.6478744190470588,0.6250127369666667} 0.64 & \cellcolor[rgb]{0.9367011412764705,0.6934798195294117,0.6531662319882353} 0.52 & \cellcolor[rgb]{0.9028614815235294,0.6299065681294118,0.6152808287} 0.65 & \cellcolor[rgb]{0.9348276152529411,0.6896369097254902,0.6504705511098039} 0.59 & \cellcolor[rgb]{0.9004151256333333,0.6254146054,0.6128478516333333} 0.69 & \cellcolor[rgb]{0.9218513348607843,0.6650340647294117,0.635032405109804} 0.68 & \cellcolor[rgb]{0.9196824685392158,0.6609281104705882,0.632461990490196} 0.56 \\
\bottomrule
\end{tabular}

}
\caption{\label{tab:overall_ranks} PRR$\uparrow$ with AlignScore aggregated over all selective QA/generation tasks and both white-box LLMs.%, totaling 35K samples. 
 The results are grouped by output length, with each interval representing approximately the same number of instances. }\end{table}
 
   \begin{table*}[t]
\centering
\footnotesize

\resizebox{\textwidth}{!}{
\begin{tabular}{l|c|c|c|c|c|c|c|c|c|c}
\toprule
\multirow{2}{*}{\textbf{UQ Method}} & \multicolumn{2}{c|}{\textbf{English, Mistral 7b}} & \multicolumn{2}{c|}{\textbf{English, Jais 13b}} & \multicolumn{2}{c|}{\textbf{Arabic, Jais 13b}} & \multicolumn{2}{c|}{\textbf{Russian, Vikhr 7b}} & \multicolumn{2}{c}{\textbf{Chinese, Yi 6b}} \\
\cline{2-11}
    & \textbf{ROC-AUC} & \textbf{PR-AUC}
    & \textbf{ROC-AUC} & \textbf{PR-AUC}
    & \textbf{ROC-AUC} & \textbf{PR-AUC}
    & \textbf{ROC-AUC} & \textbf{PR-AUC}
    & \textbf{ROC-AUC} & \textbf{PR-AUC}\\
\midrule
MSP & \cellcolor[rgb]{0.9842498738333334,0.8369886898862745,0.7783246280235294} \underline{0.65} \tiny{$\pm$ 0.03} & \cellcolor[rgb]{0.9418435698882353,0.9280538589764706,0.9201288350882354} 0.33 \tiny{$\pm$ 0.01} & \cellcolor[rgb]{0.9367011412764705,0.6934798195294117,0.6531662319882353} \underline{0.65} \tiny{$\pm$ 0.03} & \cellcolor[rgb]{0.9783266054882354,0.7990169113588235,0.7386511461764707} \underline{0.41} \tiny{$\pm$ 0.01} & \cellcolor[rgb]{0.9337138175431372,0.9321882998862745,0.9313012310098039} 0.6 \tiny{$\pm$ 0.02} & \cellcolor[rgb]{0.8620206859411765,0.9074551963235293,0.9878254853235294} 0.2 \tiny{$\pm$ 0.01} & \cellcolor[rgb]{0.9836582578333333,0.8287354144039216,0.7690800753647059} 0.59 \tiny{$\pm$ 0.02} & \cellcolor[rgb]{0.9849255762058824,0.8479150297647058,0.7906558870392157} 0.74 \tiny{$\pm$ 0.04} & \cellcolor[rgb]{0.6691882557215687,0.7264093044156863,0.9396585384392157} 0.51 \tiny{$\pm$ 0.01} & \cellcolor[rgb]{0.6817303976705882,0.7423918409254902,0.9505094434470589} 0.17 \tiny{$\pm$ 0.01} \\
Perplexity & \cellcolor[rgb]{0.9736727018,0.8973477524,0.8584952529000001} 0.62 \tiny{$\pm$ 0.02} & \cellcolor[rgb]{0.8569262456745098,0.9044285706686275,0.989693242609804} 0.29 \tiny{$\pm$ 0.01} & \cellcolor[rgb]{0.9802450306647059,0.8081382119705882,0.7477333002470589} 0.62 \tiny{$\pm$ 0.02} & \cellcolor[rgb]{0.931695915645098,0.9325418979098039,0.9338169421313726} 0.35 \tiny{$\pm$ 0.01} & \cellcolor[rgb]{0.9666105915000001,0.9077784252235295,0.8765757160705883} \underline{0.62} \tiny{$\pm$ 0.02} & \cellcolor[rgb]{0.9627817115,0.9127586490352941,0.8855681539058824} 0.23 \tiny{$\pm$ 0.01} & \cellcolor[rgb]{0.8256989195784313,0.8840607433235294,0.9979455750647059} 0.48 \tiny{$\pm$ 0.01} & \cellcolor[rgb]{0.6766845796941177,0.736034329145098,0.9462852021294117} 0.63 \tiny{$\pm$ 0.02} & \cellcolor[rgb]{0.61490285,0.649358983,0.8768415764999999} 0.5 \tiny{$\pm$ 0.01} & \cellcolor[rgb]{0.61490285,0.649358983,0.8768415764999999} 0.16 \tiny{$\pm$ 0.01} \\
Max Token Entropy & \cellcolor[rgb]{0.9841016995,0.8604220499999999,0.8061464956666666} 0.64 \tiny{$\pm$ 0.03} & \cellcolor[rgb]{0.9418435698882353,0.9280538589764706,0.9201288350882354} \underline{0.33} \tiny{$\pm$ 0.01} & \cellcolor[rgb]{0.9844470791666666,0.8397397817137255,0.7814061455764706} 0.61 \tiny{$\pm$ 0.02} & \cellcolor[rgb]{0.9797588292352941,0.8834864272549019,0.8370723575196078} 0.38 \tiny{$\pm$ 0.01} & \cellcolor[rgb]{0.8910245585529412,0.9214321063294117,0.9714899216352941} 0.58 \tiny{$\pm$ 0.02} & \cellcolor[rgb]{0.9797588292352941,0.8834864272549019,0.8370723575196078} \underline{0.24} \tiny{$\pm$ 0.01} & \cellcolor[rgb]{0.8492270432274509,0.8997249420568627,0.9922887283509805} 0.49 \tiny{$\pm$ 0.01} & \cellcolor[rgb]{0.8204138911862745,0.8803757532058823,0.9989228874411764} 0.67 \tiny{$\pm$ 0.03} & \cellcolor[rgb]{0.7285232392627451,0.797002774964706,0.9815146148450979} 0.52 \tiny{$\pm$ 0.01} & \cellcolor[rgb]{0.8336264621666667,0.8895882285,0.9964796065} 0.19 \tiny{$\pm$ 0.01} \\
PMI & \cellcolor[rgb]{0.61490285,0.649358983,0.8768415764999999} 0.44 \tiny{$\pm$ 0.01} & \cellcolor[rgb]{0.61490285,0.649358983,0.8768415764999999} 0.19 \tiny{$\pm$ 0.01} & \cellcolor[rgb]{0.61490285,0.649358983,0.8768415764999999} 0.45 \tiny{$\pm$ 0.01} & \cellcolor[rgb]{0.61490285,0.649358983,0.8768415764999999} 0.23 \tiny{$\pm$ 0.01} & \cellcolor[rgb]{0.61490285,0.649358983,0.8768415764999999} 0.47 \tiny{$\pm$ 0.01} & \cellcolor[rgb]{0.61490285,0.649358983,0.8768415764999999} 0.14 \tiny{$\pm$ 0.01} & \cellcolor[rgb]{0.61490285,0.649358983,0.8768415764999999} 0.39 \tiny{$\pm$ 0.01} & \cellcolor[rgb]{0.61490285,0.649358983,0.8768415764999999} 0.61 \tiny{$\pm$ 0.02} & \cellcolor[rgb]{0.61490285,0.649358983,0.8768415764999999} 0.5 \tiny{$\pm$ 0.01} & \cellcolor[rgb]{0.6817303976705882,0.7423918409254902,0.9505094434470589} 0.17 \tiny{$\pm$ 0.01} \\
CCP & \cellcolor[rgb]{0.852836579,0.50777808,0.575116406} \bf{0.74} \tiny{$\pm$ 0.04} & \cellcolor[rgb]{0.852836579,0.50777808,0.575116406} \bf{0.46} \tiny{$\pm$ 0.01} & \cellcolor[rgb]{0.852836579,0.50777808,0.575116406} \bf{0.68} \tiny{$\pm$ 0.03} & \cellcolor[rgb]{0.852836579,0.50777808,0.575116406} \bf{0.47} \tiny{$\pm$ 0.01} & \cellcolor[rgb]{0.852836579,0.50777808,0.575116406} \bf{0.73} \tiny{$\pm$ 0.04} & \cellcolor[rgb]{0.852836579,0.50777808,0.575116406} \bf{0.3} \tiny{$\pm$ 0.01} & \cellcolor[rgb]{0.852836579,0.50777808,0.575116406} \bf{0.67} \tiny{$\pm$ 0.03} & \cellcolor[rgb]{0.852836579,0.50777808,0.575116406} \bf{0.8} \tiny{$\pm$ 0.04} & \cellcolor[rgb]{0.852836579,0.50777808,0.575116406} \bf{0.61} \tiny{$\pm$ 0.02} & \cellcolor[rgb]{0.852836579,0.50777808,0.575116406} \bf{0.25} \tiny{$\pm$ 0.01} \\
P(True) & \cellcolor[rgb]{0.9736727018,0.8973477524,0.8584952529000001} 0.62 \tiny{$\pm$ 0.02} & \cellcolor[rgb]{0.8816813900509803,0.917546110909804,0.9778288382784314} 0.3 \tiny{$\pm$ 0.01} & \cellcolor[rgb]{0.8980319349294118,0.9243466028941176,0.9667357341529412} 0.55 \tiny{$\pm$ 0.02} & \cellcolor[rgb]{0.8042736801705882,0.8678626149117648,0.9996769126490196} 0.3 \tiny{$\pm$ 0.01} & \cellcolor[rgb]{0.7635661295607843,0.8323498010588235,0.9945323233411765} 0.53 \tiny{$\pm$ 0.01} & \cellcolor[rgb]{0.6892991246352941,0.7519281085960785,0.9568458054235294} 0.16 \tiny{$\pm$ 0.01} & \cellcolor[rgb]{0.9240202011823531,0.6691400189882354,0.6376028197294119} \underline{0.64} \tiny{$\pm$ 0.03} & \cellcolor[rgb]{0.9807976965156863,0.8111235437352942,0.7507756750235295} \underline{0.75} \tiny{$\pm$ 0.04} & \cellcolor[rgb]{0.852836579,0.50777808,0.575116406} \bf{0.61} \tiny{$\pm$ 0.02} & \cellcolor[rgb]{0.852836579,0.50777808,0.575116406} \bf{0.25} \tiny{$\pm$ 0.01} \\
\bottomrule
\end{tabular}

}

\caption{\label{tab:claim_level_multilang_results} ROC-AUC$\uparrow$ and PR-AUC$\uparrow$ (with unsupported claims as the the positive class) on the claim-level fact-checking benchmark. Warmer colors indicate better results.}
\end{table*}

 \noindent\textbf{Selective generation.}
   Tables~\ref{tab:stablelm_gen_results} and~\ref{tab:mistral_gen_results} in Appendix~\ref{sec:additional_exps} present detailed results on the selective generation task for white-box LLMs: \Stable and \Mistral v0.2, respectively. For both models, on the text summarization task, the best results are achieved by sample-diversity techniques in terms of both metrics. The majority of information-based techniques have negative or near-zero PRR in most of the cases, which indicates that they perform similarly or worse than random.
   One exception is the performance of Conditional PMI in terms of ROUGE-L-based PRR, which is drastically better than the performance of all other methods. This discrepancy might be due to limitations of the ROUGE-L metric for measuring the quality of text summarization.
   
   On MT, in terms of PRR-COMET, there is no clear winner. For Stable LM, on both datasets SAR achieves the best results, while for Mistral, basic information-based techniques are the best. However, in terms of PRR-AlignScore, the black-box techniques based on the NLI similarity are clear winners. It is worth highlighting DegMat, as in terms of PRR-AlignScore, it outperforms all techniques for all considered datasets for Stable LM and achieves the second best results for Mistral.
   
   Figure~\ref{fig:selective_generation_ranks} presents the mean PRR values for each white-box model individually, as well as the mean PRR across both models. The best methods are sample diversity techniques: SAR, DegMat, and Semantic Entropy. The best white-box techniques, SAR and Semantic Entropy, are closely competing with black-box methods. The MSP baseline trails behind in this task but still achieves reasonable performance.
 
 \noindent\textbf{Output length impact.}
   Table~\ref{tab:overall_ranks} presents the aggregated PRR scores for the two white-box LLMs across all datasets, categorized by the length of their outputs. For short generations ($< 7$ symbols), information-based methods (Maximum Probability, Perplexity, Mean Token Entropy, CCP, and TokenSAR) perform the best. Conversely, for longer outputs, sample diversity methods, especially black-box techniques based on NLI similarity, achieve superior performance. SAR, despite being inferior to black-box techniques for long outputs and slightly inferior to information-based techniques for short outputs, demonstrates the most robust results across all output lengths.

 \subsection{Fact-Checking}
   In the fact-checking evaluation, we generated biographies for 100 individuals across five languages: English (using \Mistral-v0.1 and Jais 13B), Arabic (using Jais 13B), Russian (using Vikhr 7B-0.2), and Chinese (using Yi 6B). 
      From biographies in English generated using \Mistral, we have extracted 2,499 claims; from biographies in English generated by Jais 13B -- 1,100 claims; in Arabic -- 1,031 claims; in Russian -- 3,104 claims, and in Chinese -- 2,703 claims. The percentage of claims annotated as unsupported by the automatic pipeline was 20.5\% for \Mistral-v0.1 and 16.7\% for Jais 13B in English, 17.4\% in Arabic, 58.6\% in Russian, and 20.0\% in Chinese. No more than 8\% of claims were classified as unknown across all models and languages and were discarded from the evaluation. Table~\ref{tab:claim_level_multilang_results} shows the performance of UQ techniques in fact-checking obtained using the automatic evaluation pipeline. For all tested models and languages, except Chinese, CCP consistently achieves the best performance, surpassing other methods. For Chinese, CCP and \PT yield comparable results across both evaluated metrics. This may be attributed to peculiarities of the Yi model resulting from specific fine-tuning, which enhances the model's ability to assess its own confidence.

 \subsection{Effect of Uncertainty Normalization}
 \label{sec:normalization_exp}
   
\begin{figure}[t]
     \centering
     \includegraphics[trim={0.cm 0.cm 0.cm 1.cm},clip,width=\linewidth]{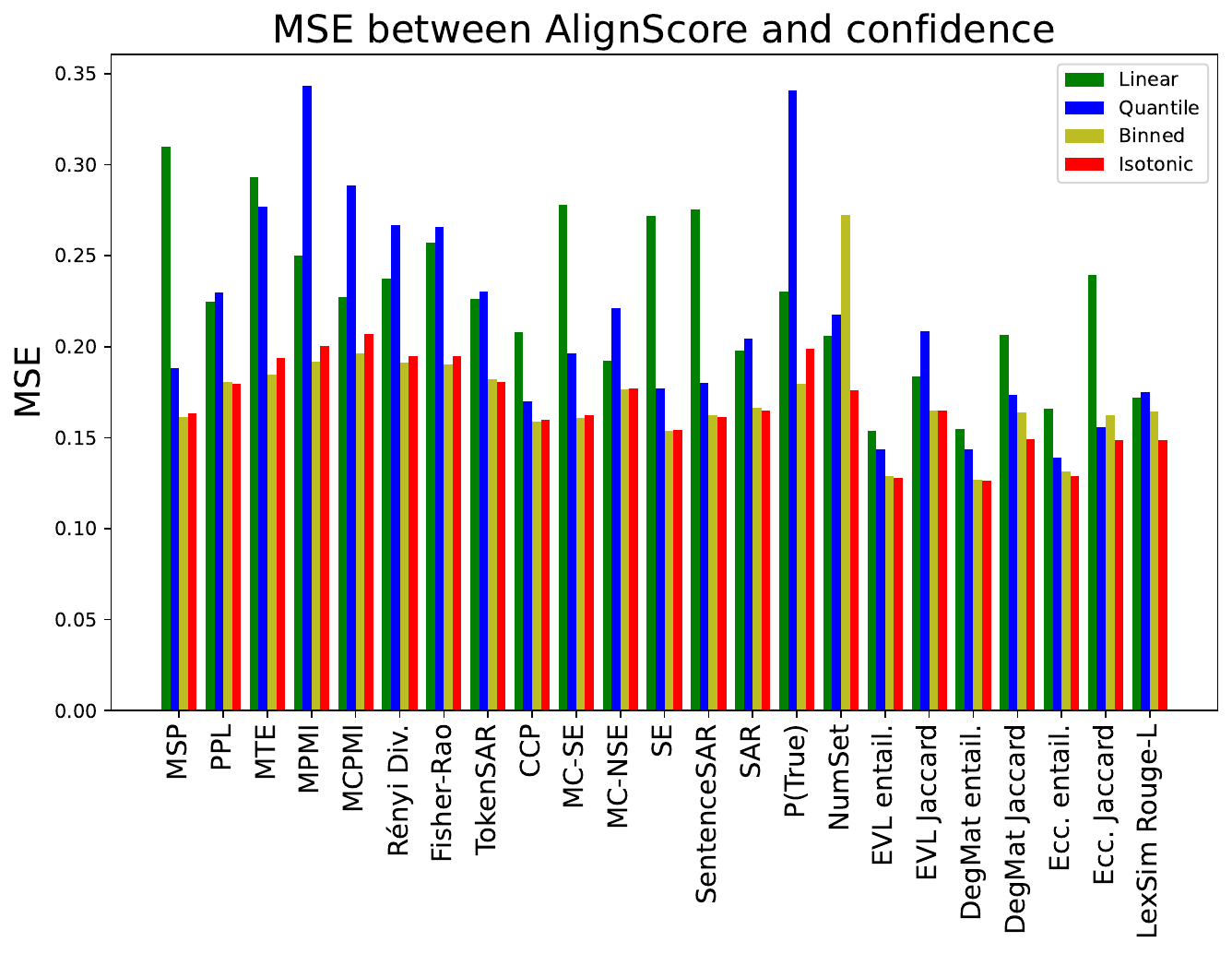}     \caption{
     MSE$\downarrow$ between AlignScore and confidence scores obtained by various normalization methods. Scorers are fitted on combined train partitions of all datasets, and MSE is averaged over their combined test partitions.
     }
     \label{fig:confidence_mse_total}
   \end{figure}

      The calibration (MSE) of confidence scores obtained by various normalization methods is presented in Figure~\ref{fig:confidence_mse_total}.
            Binned and isotonic PCC confidently outperform the linear and quantile normalization for almost all of the considered UQ techniques.
         To verify that the quality of the normalized scores does not degrade after normalization, we also evaluate their performance in selective QA/generation tasks (see Table~\ref{tab:total_confidence_prr} in Appendix~\ref{sec:normalization_appendix}). 
      We compute PRR for raw and normalized uncertainty scores and analyze the difference. Our observations indicate that scores normalized using linear, quantile, and isotonic PCC methods perform similarly to the raw scores, which is anticipated due to their monotonic properties.

 \section{Conclusion}
   In this work, we proposed a comprehensive benchmark for evaluating UQ techniques in text generation tasks. 
      The empirical investigation conducted on the developed benchmark provides several useful insights.    Overall, methods based on sample diversity perform well across selective QA/generation tasks. However, for shorter answers, we recommend using information-based methods because they often perform on par with diversity-based techniques, while introducing much less computational overhead. Specifically, for multiple-choice QA, information-based methods MSP, Perplexity, and CCP would be substantially superior. 
      For the tasks that assume longer outputs, methods based on sample diversity such as Semantic Entropy, DegMat, or Lexical Similarity are preferable. 
   It is worth highlighting that SAR consistently stands out as one of the most effective methods for short and long outputs.
   Reflexive methods in general do not demonstrate good performance. Only for big LLMs, such as GPT-4o-mini, it might be reasonable to use verbalized techniques. 
      For the fact-checking task, the best method is CCP: it demonstrates the best results and is  computationally efficient. 
   We should also note that MSP appears to be a very strong and robust baseline across all tasks and should not be discarded from the evaluation protocols. For generating human-interpretable confidence scores, we suggest normalization based on isotonic PCC as it improves confidence calibration and does not degrade performance in terms of PRR.

 \section*{Acknowledgements}
   We thank anonymous reviewers for their insightful feedback towards improving this paper. This work is supported by a grant \#848011 from the MBZUAI \& WIS Collaborative Research Program.
 
  \bibliography{custom}
 \bibliographystyle{acl_natbib}
 
 \newpage
 \clearpage
 
 \appendix
 
 \section{Detailed Description of Uncertainty Quantification Methods}
 \label{sec:appendix_methods}
   Here, we provide details of UQ methods implemented in LM-Polygraph that were omitted from the main part of the paper; see also Table~\ref{tab:ue_methods}. For the ease of notation, we will write all the uncertainty measures as functions $U(\xv)$ of an input $\xv$, though they might depend on various other instances like a generated output $\yv$.
 
 \subsection{White-Box Methods}
 \label{suppl:whitebox}
 \subsubsection{Information-Based Methods}
   \textit{Length-normalized log probability} computes the average negative log probability of generated tokens. If the score is exponentiated, it corresponds to \textit{perplexity}:
   \begin{EQA}[c]
     U_\mathrm{Perp}(\xv) = \exp\Bigl\{-\frac{1}{L} \log P(\yv \mid \xv)\Bigr\},
   \label{eq:nsp}
   \end{EQA}
   while it is convenient also to denote length-normalized sequence probability by $\bar{P}(\yv \mid \xv) = \exp\Bigl\{\frac{1}{L} \log P(\yv \mid \xv)\Bigr\}$.
   
   \textit{Mean token entropy} simply averages entropy of each token in the generated sequence:
   \begin{EQA}[c]
     U_{\HC_T}(\xv) = \frac{1}{L} \sum\nolimits_{l = 1}^L \HC(y_l \mid \yv_{<l}, \xv),
   \label{eq:entropy}
   \end{EQA}
   where $\HC(y_l \mid \yv_{<l}, \xv)$ is an entropy of the token distribution $P(y_l \mid \yv_{<l}, \xv)$.
 
   Generalizing length-normalized log probability, \textit{TokenSAR}~\cite{duan-etal-2024-shifting} computes the weighted average of the negative log probability of generated tokens based on their relevance for the entire generated text. For a given sentence similarity function $g(\cdot, \cdot)$ and token relevance function $R_T(y_k, \yv, \xv) = 1 - g(\xv \cup \yv, \xv \cup \yv \setminus y_k)$, the resulting estimate is given by the following formula:
   \begin{EQA}
     && \!\!\!\! U_\mathrm{TokenSAR}(\xv) = \mathrm{TokenSAR}(\yv, \xv) = \\
     && -\sum\nolimits_{l = 1}^L \tilde{\mathrm{R}}_T(y_l, \yv, \xv) \log P(y_l \mid \yv_{<l}, \xv),
   \end{EQA}
   where $\tilde{\mathrm{R}}_T(y_k, \yv, \xv) = \frac{\mathrm{R}_T(y_k, \yv, \xv)}{\sum\nolimits_{l = 1}^L \mathrm{R}_T(y_l, \yv, \xv)}$.
 
   \citet{takayama-arase-2019-relevant} proposed an uncertainty score based on \textit{Pointwise Mutual Information} (PMI) between generation conditioned on the prompt and unconditional generation:
      \begin{EQA}[c]
     U_\mathrm{PMI}(\xv) = \frac{1}{L} \sum\nolimits_{l = 1}^L \log \frac{P(y_l \mid \yv_{<l})}{P(y_l \mid \yv_{<l}, \xv)}.
   \end{EQA}
      \citet{van-der-poel-etal-2022-mutual} suggested a modification of this approach called \textit{Conditional Pointwise Mutual Information (CPMI)} that considers only the probabilities of those tokens, for which the entropy of the conditional distribution is above a certain threshold $\tau$:   \begin{EQA}
     && \!\!\!\! U_\mathrm{CPMI}(\xv) = -\frac{1}{L} \sum_{l = 1}^L \log P(y_l \mid \yv_{<l}, \xv) \\
     && \quad + \frac{\lambda}{L} \sum\nolimits_{l\colon \HC(y_l \mid \yv_{<l}, \xv) \ge \tau} \log P(y_l \mid \yv_{<l}),
   \end{EQA}
   where $\lambda > 0$ is another tunable parameter.
 
      \textit{Rényi divergence}~\cite{darrin-etal-2023-rainproof} computes the divergence between the probability distribution for each token and the uniform distribution: 
   \begin{EQA}
     && \!\!\!\! U_\mathrm{RD}(\xv) = \\
     && \frac{1}{L} \sum\nolimits_{l = 1}^L \frac{1}{\alpha - 1} \log \sum\nolimits\nolimits\nolimits_{i=1}^{N} \frac{P(y_i \mid \yv_{<l}, \xv)^{\alpha}}{\mathbf{q}_i^{\alpha - 1}},
   \end{EQA}
   where $\alpha > 0$ is a tunable parameter, $N$ is the number of tokens in the vocabulary, and $\mathbf{q} = \bigl[\frac{1}{N}, \dots,\frac{1}{N}\bigr]$ is a probability vector with a uniform distribution over the vocabulary.
 
   The other way to compute the distance between probability distributions is the \textit{Fisher-Rao distance}~\cite{darrin-etal-2023-rainproof}:
   \begin{EQA}
     && \!\!\!\! U_\mathrm{FR}(\xv) = \\
     && \frac{1}{L} \sum\nolimits_{l = 1}^L \frac{2}{\pi} \arccos \sum\nolimits_{i=1}^{N} \sqrt{P(y_i \mid \yv_{<l}, \xv) \cdot \mathbf{q}_i}.
   \end{EQA}

 \subsubsection{Methods Based on Sample Diversity}
   We can compute the entropy on the sequence level $\EE \bigl[-\log P(\yv \mid \xv)\bigr]$, where the expectation is taken over the sequences $\yv$ randomly generated from the distribution $P(\yv \mid \xv)$. Unfortunately, while for token level, we have an exact way of computing the entropy, for the sequence level, we need to adhere to some approximations. In practice, we can use Monte-Carlo integration, i.e. generate several sequences $\yv^{(k)}, ~ k = 1, \dots, K$ via random sampling and compute \textit{Monte Carlo Sequence Entropy}:
   \begin{EQA}[c]
     U_{\HC_S}(\xv) = -\frac{1}{K} \sum\nolimits_{k = 1}^K \log P(\yv^{(k)} \mid \xv). ~~
   \label{eq:seq_entropy}
   \end{EQA}
      We can replace $P(\yv^{(k)} \mid \xv)$ with its length-normalized version $\bar{P}(\yv^{(k)} \mid \xv)$ leading to a more reliable uncertainty measure in some cases.
 
   \textit{Semantic Entropy}~\cite{kuhn2023semantic} aims to deal with the generated sequences that have similar meanings while having different probabilities according to the model, which can significantly affect the resulting entropy value~\eqref{eq:seq_entropy}. The idea is to cluster generated sequences $\yv^{(k)}, k = 1, \dots, K$ into several semantically homogeneous clusters $\CC_m, m = 1, \dots, M$ with $M \le K$ with bi-directional entailment algorithm and average the sequence probabilities within the clusters. The resulting estimate of entropy is given by:
   \begin{EQA}[c]
     U_\mathrm{SE}(\xv) = -\sum\nolimits_{m = 1}^M \frac{|\CC_m|}{K} \log \hat{P}_m(\xv),
   \end{EQA}
   where $\hat{P}_m(\xv) = \sum_{\yv \in \CC_m} P(\yv \mid \xv)$.
 
   \textit{SentenceSAR}~\cite{duan-etal-2024-shifting} enlarges the probability of those sentences that are more relevant and convincing than others. Given sentence relevance measure $g\bigl(\yv^{(j)}, \yv^{(k)}\bigr)$ of $\yv^{(j)}$ concerning to $\yv^{(k)}$, SentenceSAR is computed as:
   \begin{EQA}
     &&\mathrm{R}_S (\yv^{(j)}, \xv) \! = \sum_{k \neq j} g\bigl(\yv^{(j)}, \yv^{(k)}\bigr) P\bigl(\yv^{(k)} \mid \xv\bigr). \\
              && U_\mathrm{SentSAR}(\xv) = \\
     && \!\!\!\! -\frac{1}{K} \sum_{k = 1}^K \log \Bigl(P(\yv^{(k)} \mid \xv) + \frac{1}{t} \mathrm{R}_S (\yv^{(k)}, \xv)\Bigr),
     \label{eq:sent_sar}
   \end{EQA}
   where $t$ is a temperature parameter used to control the scale of shifting to relevance.
 
   Combining SentenceSAR and TokenSAR results in a new method \textit{SAR}~\cite{duan-etal-2024-shifting}. In particular, in equation~\eqref{eq:sent_sar}, the generative probability $P(\yv \mid \xv)$ is replaced with the token-shifted probability $P'(\yv \mid \xv) = \exp\{-\mathrm{TokenSAR}(\yv, \xv)\}$.

 \subsubsection{Density-Based Methods}
   Let $h(\xv)$ be a latent representation of an instance $\xv$. The \textit{Mahalanobis Distance} (MD; \citet{lee2018simple}) method fits a Gaussian centered at the training data centroid $\mu$ with an empirical covariance matrix $\Sigma$. The uncertainty score is the Mahalanobis distance between $h(\xv)$ and $\mu$:
   \begin{EQA}[c]
     U_\mathrm{MD}(\xv) = \bigl(h(\xv) - \mu\bigr)^{T} \Sigma^{-1} \bigl(h(\xv) - \mu\bigr).
   \end{EQA}
      We suggest using the last hidden state of the encoder averaged over non-padding tokens or the last hidden state of the decoder averaged over all generated tokens as $h(\xv)$.
   
   The \textit{Robust Density Estimation} (RDE; \citet{yoo-etal-2022-detection}) method improves over MD by reducing the dimensionality of $h(\xv)$ via the PCA decomposition. Additionally, the covariance matrix $\Sigma$ for each class is computed using the Minimum Covariance Determinant estimation method~\cite{Rousseeuw84leastmedian}. The uncertainty score is computed as the Mahalanobis distance but in the space of reduced dimensionality.
 
   \citet{ren2023outofdistribution} showed that it might be useful to adjust the Mahalanobis distance score by subtracting from it the other Mahalanobis distance $\mathrm{MD}_0(\xv)$ computed for some large general-purpose dataset covering many domains (e.g. C4~\cite{raffel2020exploring}). The resulting \textit{Relative Mahalanobis Distance} score is
   \begin{EQA}[c]
     U_\mathrm{RMD}(\xv) = \mathrm{MD}(\xv) - \mathrm{MD}_0(\xv).
   \end{EQA}

 \subsection{Black-Box Methods}
 \label{suppl:blackbox}
 
 \subsubsection{Methods Based on Sample Diversity}
   Sample diversity methods sample multiple predictions from a LLM for the same prompt and analyze the diversity of the outputs across different samples. The idea is that if the LLM consistently outputs similar answers, it is confident, whereas varying outputs indicate high uncertainty. Since the LLM might output the same meaning in various surface forms by rephrasing its answers, the approaches from this category usually construct a matrix $S=(s_{ij})$ representing similarities between responses based on some semantic similarity measure and then cluster the responses into groups of answers with the same meanings.
   
   Following~\citet{lin2023generating}, we consider two similarity measures for responses. The first one is the Jaccard similarity 
   $     s(\yv, \yv') = |\yv \cap \yv'| / |\yv \cup \yv'|,
   $   where the sequences $\yv$ and $\yv'$ are considered just as sets of words.
   Another similarity measure is based on Natural Language Inference (NLI). For each pair of input sequences, an NLI model provides two probabilities: $\hat{p}_{\mathrm{entail}}(\yv, \yv')$ -- the degree of entailment between the sequences and $\hat{p}_{\mathrm{contra}}(\yv, \yv')$ -- the degree the contradiction between them. The similarity between sequences $\yv$ and $\yv'$ is computed as $s_{\mathrm{entail}}(\yv, \yv') = \hat{p}_{\mathrm{entail}}(\yv, \yv')$ or $s_{\mathrm{contra}}(\yv, \yv') = 1 - \hat{p}_{\mathrm{contra}}(\yv, \yv')$. 
      Following~\citet{kuhn2023semantic}, we use the DeBERTa-large NLI model~\cite{he2021deberta}.
 
   One of the simplest techniques that leverages the idea of meaning diversity for UQ is \textit{Number of Semantic Sets}. We adopt an iterative approach by sequentially examining responses from the first to the last while making pairwise comparisons between them (each pair has indexes $j_1$ and $j_2$, $j_2 > j_1$). The number of semantic sets initially equals the total number of generated answers $K$. If the condition $\hat{p}_{\mathrm{entail}}(\yv_{j_1},\yv_{j_2}) > \hat{p}_{\mathrm{contra}}(\yv_{j_1},\yv_{j_2}) $ and $\hat{p}_{\mathrm{entail}}(\yv_{j_2},\yv_{j_1}) > \hat{p}_{\mathrm{contra}}(\yv_{j_2}, \yv_{j_1}) $ is fulfilled we put this two sentences into one cluster. The computation is done for all the pairs of answers, and then the resulting number of distinct sets $U_{NumSemSets}$ is reported.
      This measure is simple yet it has many limitations. It can only take integer values and it assumes that the semantic equivalences derived from the NLI model are always transitive. 
 
   \textit{Sum of Eigenvalues of the Graph Laplacian}~\cite{lin2023generating} represents a more advanced approach from this category.
      Let's consider a similarity matrix $S_{j_1 j_2} = \bigl(s(\yv_{j_1}, \yv_{j_2}) + s(\yv_{j_2}, \yv_{j_1})\bigr) / 2$. Averaging is done to obtain better consistency. The Laplacian for matrix $S$ is given by the following formula $L = I - D^{-\frac{1}{2}} S  D^{-\frac{1}{2}}$, where $D$ is a diagonal matrix and $D_{ii} = \sum_{j = 1}^K{S_{ij}}$. Consequently, the following formula is derived: $U_{EigV} = \sum_{k = 1}^{K}{\max(0, 1 - \lambda_{k})}$, where $\lambda_k$ are the eigenvalues of matrix $L$. This value is a continuous analogue of $U_{NumSemSets}$. 
     
   $U_{EigV}$ and $U_{NumSemSets}$ have a common disadvantage: they can not provide uncertainty for each answer. \citet{lin2023generating} demonstrates that we can extract it from the diagonal \textit{Degree Matrix} $D$ computed above. The idea is that 
       elements on the diagonal of $D$ are sums of similarities between the given answer and all other answers, and the corrected trace of $D$ provides an average pairwise distance between answers. The larger the pairwise distance is the higher is uncertainty: $U_{Deg} = 1 - trace(D) / K^{2}$.
 
   A drawback of previously considered methods is the limited knowledge of the actual embedding space for the different answers since we only have measures of their similarities. Nevertheless, we can overcome this limitation by taking advantage of the inferential capabilities of the graph Laplacian, which makes it easier to obtain the coordinates of the answers. Let us introduce $\uv_1, \dots, \uv_{k} \in R^{K}$ as the eigenvectors of $L$ that correspond to $k$ smallest eigenvalues. We can efficiently construct an informative embedding $\vv_{j} = [\uv_{1,j}, \dots, \uv_{k,j}]$ for an answer $\yv_{j}$. \citet{lin2023generating} suggest
   \textit{Eccentricity} as uncertainty score -- the average distance from the center in the space of constructed embeddings:
      $U_{Ecc} = \bigl\|[\tilde{\vv}_{1}^{T}, \dots, \tilde{\vv}_{K}^{T}]\bigr\|_{2}$, where $\tilde{\vv}_{j} = \vv_{j} - \frac{1}{K}\sum_{\ell = 1}^{K}{\vv_{\ell}}$.
   
   \textit{Lexical Similarity} is a measure proposed by~\cite{fomicheva-etal-2020-unsupervised} that computes how similar two words or phrases are in terms of their meaning. Since the original article is dedicated to machine translation, this measure calculates the average similarity score between all pairs of translation hypotheses in a set, using a similarity measure based on the overlap of their lexical items. Different metrics can be used, such as ROUGE-1, ROUGE-2, ROUGE-L, and BLEU. For our task, this measure iterates over all responses and calculates the average score with other answers. 
 
   \subsubsection{Empirical Approximations of White-box Methods}
   Following~\cite{tian-etal-2023-just}, we introduce several empirical variations of white-box methods that can be computed in a black-box setting. \textit{LabelProb} is a black-box approximation of MSP. Given $K$ sampled outputs from the model, we can estimate the model-assigned probability for each of the outputs $\hat{\mathrm{P}}\bigl(\yv^{(j)} \mid \xv\bigr)$, based on its relative frequency among the samples:
        $\hat{\mathrm{P}}(\yv^{(j)} \mid \xv) = \frac{1}{K}\sum_{i=1}^{K}\mathbb{I}\bigl(\yv^{(i)} = \yv^{(j)}\bigr).$
      LabelProb is computed by considering the probability of the most likely sample:
   \begin{EQA}[c]
     U_\mathrm{LabelProb}(\xv) = 1 - \max_{j}\hat{\mathrm{P}}\bigl(\yv^{(j)} \mid \xv\bigr).
   \end{EQA}
 
   Similarly, it is possible to estimate the black-box equivalent of \textit{Semantic Entropy} by calculating semantic cluster probabilities based on the relative frequencies of samples within these clusters: $\hat{P}_m(\xv) = \sum_{\yv \in \CC_m} \hat{P}(\yv \mid \xv).$
 
   We also implement a black-box approximation of \PT~\cite{kadavath2022language}, where the model is repeatedly queried for its confidence, and the relative frequency of the ``True'' output is taken as the confidence measure:
   \begin{EQA}
     && \hat{\mathrm{P}}(\text{``True''} \mid \xv) = \frac{1}{K}\sum_{i=1}^{K}\mathbb{I}(\yv^{(i)} = \text{``True''}),
     \\
     &&
     U_{\mathrm{PTrue}_{BB}}(\xv) = 1 - \hat{\mathrm{P}}(\text{``True''} \mid \xv).
   \end{EQA}

 \subsubsection{Reflexive Methods}
   Instruction-tuned LLMs can also be directly prompted to output a level of confidence as a part of their output, as shown by~\citet{tian-etal-2023-just}. 
    
   \textit{Linguistic 1S} prompts the model to output its confidence along with the answer by selecting it from the list of predetermined linguistic expressions of confidence. The selected answer is mapped to a floating-point confidence level following~\cite{Fagen2023Perception}.
 
   \textit{Verbalized 1S Top1/TopK} prompts the model to generate both the answer and its confidence in a single output, with the confidence expressed directly as a floating-point number. Top1/TopK approaches specify how many guesses (with corresponding confidences) the model is asked to output. 
 
   \textit{Verbalized 2S Top1/TopK} differs from the previous approach by separating the answer and confidence estimation into two distinct turns of interaction with a LLM. 
       
   \textit{Verbalized 2S CoT} asks the model to reason about the question and output its answer in the first response and numerical confidence in the second.

 \subsection{Claim-Level Methods}
 \label{app:claim_level_methods}
   While the aforementioned methods operate with the entire generated sequence, it is often desirable to estimate uncertainty for individual claims to pinpoint hallucinations within the generated text. Suppose $C$ denotes a set of token indices corresponding to a particular claim. Many UQ methods can be straightforwardly adopted for the claim level by simply considering only a subset of tokens corresponding to the set $C$ instead of all tokens in the sequence. We consider claim-level generalization for MSP, Mean Token Entropy, Perplexity, and PMI. 
 
   \textit{\PT}, adapting the approach from~\cite{kadavath2022language} to the claim level, quantifies claim uncertainty by prompting the LLM to assess the truthfulness of each generated claim: 
      \begin{EQA}[c]
     U_{\mathrm{P_\text{True}}}(C \mid \xv) = 1 - P(y_1 = \text{``True''} \mid C, \xv).
   \end{EQA}
 
   \textit{Claim-Conditioned Probability} (CCP)   
   quantifies uncertainty by evaluating the semantic similarity between the original claim and perturbed versions where each token is replaced with its alternative generations. CCP utilizes a Natural Language Inference (NLI) model to compare the original claim $y_{i \in C, i \le j}$ with variations where token $y_j$ is replaced with top-$K$ alternatives $y^k_j$ from the model's output distribution:
   \begin{EQA}
     && \!\!\!\! \mathrm{CCP}(y_j \mid \yv_{<j}, \xv) = \\
     && \dfrac{\sum_{k: \texttt{NLI}(y^k_j, y_j) = \texttt{`e'}} P(y^k_j \mid \yv_{<j}, \xv)} {\sum_{k: \texttt{NLI}(y^k_j, y_j) \in \left\{ \texttt{`e'}, \texttt{`c'}\right\}} P(y^k_j \mid \yv_{<j} , \xv)},
   \end{EQA}
      The resulting uncertainty measure becomes
   \begin{EQA}[c]
     U_\mathrm{CCP}(C \mid \xv) = 1 - \prod_{j \in C} \mathrm{CCP}(y_j \mid \yv_{<j}, \xv).
   \end{EQA}
      Here, $\texttt{NLI}(y^k_j, y_j) = \texttt{`e'}$ denotes that the NLI model predicts an entailment relation between the original claim and the modified claim where $y_j$ is replaced with $y^k_j$. CCP effectively measures the proportion of high-probability token alternatives that preserve the original claim's semantic meaning according to the NLI model.

 \section{Detailed Description of Uncertainty Normalization Methods}
 \label{sec:normalization_theory}
   \textit{Linear scaling} first computes confidence scores by negating uncertainty scores in the calibration set: $c_i = -u_i$. Then, for a new model output with a corresponding uncertainty score $u$ and confidence $c(u) = - u$, the normalized confidence is
   $c_s(u) = (c(u) - \min_i c_i) / (\max_i c_i - \min_i c_i)$.
      To ensure the uncertainty scores for tested instances remain within the $[0, 1]$ interval, the confidence scores are clipped accordingly.
 
   \textit{Quantile scaling} computes confidence using the uncertainty cumulative distribution function estimated using calibration data: $c(u) = 1 - \frac{1}{N}\sum\nolimits_{i=1}^{N}\mathbb{I}(u_i \leq u)$,
   where $N$ is the number of data points in $\mathcal{D}_{calib}$ and $u_i \in \mathcal{D}_{calib}$. 
      This approach naturally bounds the confidence between 0 and 1 and does this with consideration to the distribution of uncertainty values in the calibration set.
 
   \textit{Binned PCC} splits the calibration set into non-intersecting bins based on uncertainty values $u_i$. Thus, a bin is a set of indices $\mathcal{B}^j = \{i\colon b^j_{\min} \le u_i < b^j_{\max}\}$, where $b^j_{\min}$ and $b^j_{\max}$ are left and right boundaries of the $j$-th bin respectively.
 
   Then, for a new data point with a raw uncertainty score $u$, a calibrated confidence score is $c(u) = \frac{1}{|\mathcal{B}'|} \sum\nolimits_{i \in \mathcal{B}'}q_i$,
   where $\mathcal{B}'$ is the calibration bin, for which the following holds: $b'_{\min} \le u < b'_{\max}$.
   
   To remedy this problem, we propose \textit{Isotonic PCC}. In this method the confidence score is obtained as $c(u) = \mathrm{CIR}(u)$, where CIR is a fitted Centered Isotonic Regression that predicts the quality of the response based on its raw uncertainty score, maintaining strict monotonicity, i.e. higher uncertainty always produces lower confidence.

 \onecolumn
 
 \newpage
 \clearpage

 \section{Detailed Experimental Results for Uncertainty Quantification Methods}
 \label{sec:additional_exps}
    
   \begin{table*}[!ht] \centering\resizebox{0.75\textwidth}{!}{
\begin{tabular}{l|c|c|c|c|c|c}
\toprule
\multirow{2}{*}{\textbf{UQ Method}} & \multicolumn{1}{c|}{\textbf{CoQA}} & \multicolumn{1}{c|}{\textbf{TriviaQA}} & \multicolumn{1}{c|}{\textbf{MMLU}} & \multicolumn{1}{c|}{\textbf{GSM8k}}  & \multirow{2}{*}{\textbf{Mean Rank}}& \multirow{2}{*}{\textbf{Mean PRR}} \\ \cline{2-5}
    & \textbf{AlignScore} & \textbf{AlignScore}& \textbf{Accuracy} & \textbf{Accuracy} & \\\midrule
Maximum Sequence Probability & \cellcolor[rgb]{0.9575785619705882,0.7384632635882353,0.686089707382353} 0.28\tiny{$\pm$0.01} & \cellcolor[rgb]{0.9102005491941176,0.643382456317647,0.6225797599} 0.56\tiny{$\pm$0.01} & \cellcolor[rgb]{0.852836579,0.50777808,0.575116406} \textbf{0.52\tiny{$\pm$0.01}} & \cellcolor[rgb]{0.9544540133274511,0.7312163185843138,0.6804751970764706} 0.35\tiny{$\pm$0.01} & \cellcolor[rgb]{0.9404481933235294,0.7011656391372549,0.658557593745098} 8.88 & \cellcolor[rgb]{0.852836579,0.50777808,0.575116406} \textbf{0.43} \\
Perplexity & \cellcolor[rgb]{0.9796923648137255,0.8051528802058824,0.7446909254705882} 0.24\tiny{$\pm$0.01} & \cellcolor[rgb]{0.9175136022176471,0.6568221562117647,0.6298915758705882} 0.55\tiny{$\pm$0.01} & \cellcolor[rgb]{0.9385746672999999,0.6973227293333333,0.6558619128666667} 0.41\tiny{$\pm$0.01} & \cellcolor[rgb]{0.9077541933039216,0.6388904935882354,0.6201467828333334} 0.43\tiny{$\pm$0.01} & \cellcolor[rgb]{0.9497671903627452,0.7203459010784313,0.6720534316176471} 9.25 & \cellcolor[rgb]{0.8790560841823529,0.5840607685176471,0.5944135352882353} 0.41 \\
\bottomrule
\end{tabular}
}\caption{\label{tab:stablelm_cls_results} PRR$\uparrow$ 50\% with various generation quality metrics for UQ methods in selective QA tasks with the \Stable model. Warmer color indicates better results.}\end{table*}
 
   \begin{table*}[!ht] \centering\resizebox{0.75\textwidth}{!}{\begin{tabular}{l|c|c|c|c|c|c}
\toprule
\multirow{2}{*}{\textbf{UQ Method}} & \multicolumn{1}{c|}{\textbf{CoQA}} & \multicolumn{1}{c|}{\textbf{TriviaQA}} & \multicolumn{1}{c|}{\textbf{MMLU}} & \multicolumn{1}{c|}{\textbf{GSM8k}}  & \multirow{2}{*}{\textbf{Mean Rank}} & \multirow{2}{*}{\textbf{Mean PRR}} \\ \cline{2-5}
    & \textbf{AlignScore} & \textbf{AlignScore}& \textbf{Accuracy} & \textbf{Accuracy} & \\\midrule
Maximum Sequence Probability & \cellcolor[rgb]{0.8844643114450981,0.5946066788549021,0.5989826965745099} \underline{0.33\tiny{$\pm$0.01}} & \cellcolor[rgb]{0.8817601978137255,0.5893337236862746,0.5966981159313726} 0.63\tiny{$\pm$0.01} & \cellcolor[rgb]{0.864597965917647,0.5433394684235294,0.5836202017019608} \underline{0.47\tiny{$\pm$0.01}} & \cellcolor[rgb]{0.9004151256333333,0.6254146054,0.6128478516333333} 0.47\tiny{$\pm$0.01} & \cellcolor[rgb]{0.9028614815235294,0.6299065681294118,0.6152808287} \underline{6.38} & \cellcolor[rgb]{0.852836579,0.50777808,0.575116406} \textbf{0.48} \\
Perplexity & \cellcolor[rgb]{0.9367011412764705,0.6934798195294117,0.6531662319882353} 0.29\tiny{$\pm$0.01} & \cellcolor[rgb]{0.8734190061058824,0.5700105097411765,0.5899980484784314} 0.64\tiny{$\pm$0.01} & \cellcolor[rgb]{0.864597965917647,0.5433394684235294,0.5836202017019608} \underline{0.47\tiny{$\pm$0.01}} & \cellcolor[rgb]{0.9791396989627451,0.8021675484411765,0.7416485506941177} 0.31\tiny{$\pm$0.01} & \cellcolor[rgb]{0.9819030282176471,0.8170942072647058,0.7568604245764706} 10.62 & \cellcolor[rgb]{0.9077541933039216,0.6388904935882354,0.6201467828333334} 0.43 \\
\bottomrule
\end{tabular}
}\caption{\label{tab:mistral_cls_results} PRR$\uparrow$ 50\% with various generation quality metrics for UQ methods in selective QA tasks with the \Mistral v0.2 model. Warmer color indicates better results.}\end{table*}
 
   \begin{table*}[!ht] 
\centering
\resizebox{0.62\textwidth}{!}{\begin{tabular}{l|c|c|c|c|c}
\toprule
\multirow{2}{*}{\textbf{UQ Method}} & \multicolumn{1}{c|}{\textbf{CoQA}} & \multicolumn{1}{c|}{\textbf{TriviaQA}} & \multicolumn{1}{c|}{\textbf{MMLU}} & \multirow{2}{*}{\textbf{Mean Rank}} & \multirow{2}{*}{\textbf{Mean PRR}} \\ \cline{2-4}
    & \textbf{AlignScore} & \textbf{AlignScore}& \textbf{Accuracy} & \\\midrule
NumSet & \cellcolor[rgb]{0.953077067017647,0.9210455325882353,0.9030752965411765} 0.20\tiny{$\pm$0.01} & \cellcolor[rgb]{0.9575785619705882,0.7384632635882353,0.686089707382353} 0.54\tiny{$\pm$0.01} & \cellcolor[rgb]{0.8015810339588235,0.8657637386764706,0.9997826392686274} 0.29\tiny{$\pm$0.00} & \cellcolor[rgb]{0.8256989195784313,0.8840607433235294,0.9979455750647059} 13.17 & \cellcolor[rgb]{0.9781854635254902,0.8875721630666666,0.8432079741549019} 0.34 \\
EigValLaplacian NLI Score Entail. & \cellcolor[rgb]{0.8925766523392157,0.6104255443607843,0.6058364385039215} \underline{0.34\tiny{$\pm$0.01}} & \cellcolor[rgb]{0.852836579,0.50777808,0.575116406} \textbf{0.60\tiny{$\pm$0.00}} & \cellcolor[rgb]{0.8933603506784313,0.9224036051843137,0.9699051924745098} 0.32\tiny{$\pm$0.00} & \cellcolor[rgb]{0.8952807659705883,0.6156984995294118,0.6081210191470588} \underline{4.50} & \cellcolor[rgb]{0.852836579,0.50777808,0.575116406} \textbf{0.42} \\
EigValLaplacian Jaccard Score & \cellcolor[rgb]{0.8925766523392157,0.6104255443607843,0.6058364385039215} \underline{0.34\tiny{$\pm$0.01}} & \cellcolor[rgb]{0.8952807659705883,0.6156984995294118,0.6081210191470588} 0.58\tiny{$\pm$0.00} & \cellcolor[rgb]{0.8015810339588235,0.8657637386764706,0.9997826392686274} 0.29\tiny{$\pm$0.00} & \cellcolor[rgb]{0.9841016995,0.8604220499999999,0.8061464956666666} 8.33 & \cellcolor[rgb]{0.9175136022176471,0.6568221562117647,0.6298915758705882} 0.40 \\
\bottomrule
\end{tabular}
}\caption{\label{tab:gpt4o_cls_results} PRR$\uparrow$ 50\% with various generation quality metrics for black-box UQ methods in selective QA tasks with the GPT-4o-mini model. Warmer color indicates better results.}\end{table*}

   \begin{table*}[!ht] 
\centering
\resizebox{0.62\textwidth}{!}{\begin{tabular}{l|c|c|c|c|c}
\toprule
\multirow{2}{*}{\textbf{UQ Method}} & \multicolumn{1}{c|}{\textbf{CoQA}} & \multicolumn{1}{c|}{\textbf{TriviaQA}} & \multicolumn{1}{c|}{\textbf{MMLU}} & \multirow{2}{*}{\textbf{Mean Rank}} & \multirow{2}{*}{\textbf{Mean PRR}} \\ \cline{2-4}
    & \textbf{AlignScore} & \textbf{AlignScore}& \textbf{Accuracy} & \\\midrule
NumSet & \cellcolor[rgb]{0.9849255762058824,0.8479150297647058,0.7906558870392157} 0.23\tiny{$\pm$0.01} & \cellcolor[rgb]{0.8790560841823529,0.5840607685176471,0.5944135352882353} \underline{0.56\tiny{$\pm$0.01}} & \cellcolor[rgb]{0.9683898066058824,0.766374750054902,0.7090466697431372} 0.19\tiny{$\pm$0.01} & \cellcolor[rgb]{0.9844312501823529,0.8554192419058824,0.7999502522156863} 6.33 & \cellcolor[rgb]{0.9441952453705882,0.708851458745098,0.6639489555019608} 0.33 \\
EigValLaplacian NLI Score Entail. & \cellcolor[rgb]{0.852836579,0.50777808,0.575116406} \textbf{0.34\tiny{$\pm$0.01}} & \cellcolor[rgb]{0.852836579,0.50777808,0.575116406} \textbf{0.58\tiny{$\pm$0.01}} & \cellcolor[rgb]{0.852836579,0.50777808,0.575116406} \textbf{0.23\tiny{$\pm$0.01}} & \cellcolor[rgb]{0.852836579,0.50777808,0.575116406} \textbf{1.33} & \cellcolor[rgb]{0.852836579,0.50777808,0.575116406} \textbf{0.38} \\
EigValLaplacian Jaccard Score & \cellcolor[rgb]{0.9423217193470588,0.7050085489411765,0.6612532746235295} 0.29\tiny{$\pm$0.01} & \cellcolor[rgb]{0.9150932609745098,0.6523663817764705,0.6274457140333334} 0.53\tiny{$\pm$0.01} & \cellcolor[rgb]{0.6449978024117646,0.6934178380941176,0.9144631795921568} 0.05\tiny{$\pm$0.01} & \cellcolor[rgb]{0.876805309,0.9151164254999999,0.9804355785000001} 10.50 & \cellcolor[rgb]{0.9807976965156863,0.8111235437352942,0.7507756750235295} 0.29 \\
\bottomrule
\end{tabular}
}\caption{\label{tab:stablelm_instruct_cls_results} PRR$\uparrow$ 50\% with various generation quality metrics for black-box UQ methods in selective QA tasks with the \Stable Chat model. Warmer color indicates better results.}\end{table*}
 
   \begin{table*}[!ht] 
\centering
\resizebox{0.62\textwidth}{!}{\begin{tabular}{l|c|c|c|c|c}
\toprule
\multirow{2}{*}{\textbf{UQ Method}} & \multicolumn{1}{c|}{\textbf{CoQA}} & \multicolumn{1}{c|}{\textbf{TriviaQA}} & \multicolumn{1}{c|}{\textbf{MMLU}} & \multirow{2}{*}{\textbf{Mean Rank}} & \multirow{2}{*}{\textbf{Mean PRR}} \\ \cline{2-4}
    & \textbf{AlignScore} & \textbf{AlignScore}& \textbf{Accuracy} & \\\midrule
NumSet & \cellcolor[rgb]{0.9802450306647059,0.8081382119705882,0.7477333002470589} 0.21\tiny{$\pm$0.01} & \cellcolor[rgb]{0.9497671903627452,0.7203459010784313,0.6720534316176471} 0.51\tiny{$\pm$0.01} & \cellcolor[rgb]{0.61490285,0.649358983,0.8768415764999999} 0.20\tiny{$\pm$0.01} & \cellcolor[rgb]{0.9196757213862745,0.930583412827451,0.9472468817078432} 10.50 & \cellcolor[rgb]{0.9830083599196078,0.8230648707941176,0.7629451741294118} 0.31 \\
EigValLaplacian NLI Score Entail. & \cellcolor[rgb]{0.9175136022176471,0.6568221562117647,0.6298915758705882} 0.26\tiny{$\pm$0.01} & \cellcolor[rgb]{0.852836579,0.50777808,0.575116406} \textbf{0.54\tiny{$\pm$0.01}} & \cellcolor[rgb]{0.9337138175431372,0.9321882998862745,0.9313012310098039} 0.27\tiny{$\pm$0.01} & \cellcolor[rgb]{0.852836579,0.50777808,0.575116406} \textbf{3.50} & \cellcolor[rgb]{0.852836579,0.50777808,0.575116406} \textbf{0.36} \\
EigValLaplacian Jaccard Score & \cellcolor[rgb]{0.9845960255235294,0.8529178378588236,0.7968521304901961} 0.19\tiny{$\pm$0.01} & \cellcolor[rgb]{0.8569262456745098,0.9044285706686275,0.989693242609804} 0.42\tiny{$\pm$0.01} & \cellcolor[rgb]{0.6569731755647059,0.7100258308470588,0.927496270517647} 0.21\tiny{$\pm$0.01} & \cellcolor[rgb]{0.6996157421568627,0.7642642360784314,0.9642295513921568} 15.00 & \cellcolor[rgb]{0.9337138175431372,0.9321882998862745,0.9313012310098039} 0.27 \\
\bottomrule
\end{tabular}
}\caption{\label{tab:mistral_instruct_cls_results} PRR$\uparrow$ 50\% with various generation quality metrics for black-box UQ methods in selective QA tasks with the \Mistral v0.2 Instruct model. Warmer color indicates better results.}\end{table*}
 
   \begin{table*}[!ht] 
\centering
\resizebox{0.9\textwidth}{!}{\begin{tabular}{l|c|c|c|c|c|c|c|c}
\toprule
\multirow{2}{*}{\textbf{UQ Method}} & \multicolumn{2}{c|}{\textbf{XSUM}} & \multicolumn{2}{c|}{\textbf{WMT14 Fr-En}} & \multicolumn{2}{c|}{\textbf{WMT19 De-En}}& \multirow{2}{*}{\textbf{Mean Rank}}& \multirow{2}{*}{\textbf{Mean PRR}}\\ \cline{2-7}
    & \textbf{ROUGE-L} & \textbf{AlignScore}& \textbf{COMET} & \textbf{AlignScore}& \textbf{COMET} & \textbf{AlignScore}& \\\midrule
Maximum Sequence Probability & \cellcolor[rgb]{0.61490285,0.649358983,0.8768415764999999} -0.24\tiny{$\pm$0.01} & \cellcolor[rgb]{0.9613407263117647,0.9142840690588235,0.8885271948941176} -0.09\tiny{$\pm$0.00} & \cellcolor[rgb]{0.9513294646843138,0.7239693735803922,0.6748606867705882} 0.34\tiny{$\pm$0.01} & \cellcolor[rgb]{0.9337138175431372,0.9321882998862745,0.9313012310098039} 0.09\tiny{$\pm$0.01} & \cellcolor[rgb]{0.852836579,0.50777808,0.575116406} \textbf{0.50\tiny{$\pm$0.00}} & \cellcolor[rgb]{0.9337138175431372,0.9321882998862745,0.9313012310098039} 0.11\tiny{$\pm$0.01} & \cellcolor[rgb]{0.9790880158705882,0.8856170452000001,0.8401505184058824} 12.67 & \cellcolor[rgb]{0.9790880158705882,0.8856170452000001,0.8401505184058824} 0.12 \\
Perplexity & \cellcolor[rgb]{0.9418435698882353,0.9280538589764706,0.9201288350882354} 0.20\tiny{$\pm$0.01} & \cellcolor[rgb]{0.9849255762058824,0.8479150297647058,0.7906558870392157} 0.02\tiny{$\pm$0.00} & \cellcolor[rgb]{0.9683898066058824,0.766374750054902,0.7090466697431372} 0.31\tiny{$\pm$0.01} & \cellcolor[rgb]{0.8620206859411765,0.9074551963235293,0.9878254853235294} 0.06\tiny{$\pm$0.01} & \cellcolor[rgb]{0.9544540133274511,0.7312163185843138,0.6804751970764706} 0.38\tiny{$\pm$0.00} & \cellcolor[rgb]{0.876805309,0.9151164254999999,0.9804355785000001} 0.07\tiny{$\pm$0.01} & \cellcolor[rgb]{0.9497716033000001,0.923750118,0.9088945372} 14.58 & \cellcolor[rgb]{0.9774267028058823,0.7958111725039216,0.7356687317450981} 0.17 \\
\bottomrule
\end{tabular}
}\caption{\label{tab:stablelm_gen_results} PRR$\uparrow$ 50\% with various generation quality metrics for UQ methods in selective generation tasks with the \Stable model. Warmer color indicates better results.}\end{table*}
 
   \begin{table*}[!ht] 
\centering
\resizebox{0.9\textwidth}{!}{\begin{tabular}{l|c|c|c|c|c|c|c|c}
\toprule
\multirow{2}{*}{\textbf{UQ Method}} & \multicolumn{2}{c|}{\textbf{XSUM}} & \multicolumn{2}{c|}{\textbf{WMT14 Fr-En}} & \multicolumn{2}{c|}{\textbf{WMT19 De-En}}& \multirow{2}{*}{\textbf{Mean Rank}}& \multirow{2}{*}{\textbf{Mean PRR}} \\ \cline{2-7}
    & \textbf{ROUGE-L} & \textbf{AlignScore}& \textbf{COMET} & \textbf{AlignScore}& \textbf{COMET} & \textbf{AlignScore}& \\\midrule
Maximum Sequence Probability & \cellcolor[rgb]{0.9276891842254902,0.9318890695490196,0.9382935886568627} 0.06\tiny{$\pm$0.00} & \cellcolor[rgb]{0.9797588292352941,0.8834864272549019,0.8370723575196078} 0.06\tiny{$\pm$0.00} & \cellcolor[rgb]{0.9646785691470589,0.756126767372549,0.7003363715784314} 0.31\tiny{$\pm$0.02} & \cellcolor[rgb]{0.915574114,0.9297565972666666,0.9515550793333334} 0.08\tiny{$\pm$0.01} & \cellcolor[rgb]{0.9102005491941176,0.643382456317647,0.6225797599} 0.45\tiny{$\pm$0.01} & \cellcolor[rgb]{0.9459084460607843,0.9259866385215686,0.914542637127451} 0.12\tiny{$\pm$0.01} & \cellcolor[rgb]{0.9797588292352941,0.8834864272549019,0.8370723575196078} 12.08 & \cellcolor[rgb]{0.9756268974411765,0.7893996947941176,0.729703902882353} 0.18 \\
Perplexity & \cellcolor[rgb]{0.61490285,0.649358983,0.8768415764999999} -0.19\tiny{$\pm$0.00} & \cellcolor[rgb]{0.8204138911862745,0.8803757532058823,0.9989228874411764} -0.15\tiny{$\pm$0.00} & \cellcolor[rgb]{0.9126469050843138,0.6478744190470588,0.6250127369666667} \underline{0.37\tiny{$\pm$0.02}} & \cellcolor[rgb]{0.9736727018,0.8973477524,0.8584952529000001} 0.12\tiny{$\pm$0.01} & \cellcolor[rgb]{0.8734190061058824,0.5700105097411765,0.5899980484784314} \underline{0.49\tiny{$\pm$0.01}} & \cellcolor[rgb]{0.9678868848333333,0.9061183506196079,0.873578236792157} 0.14\tiny{$\pm$0.01} & \cellcolor[rgb]{0.9797588292352941,0.8834864272549019,0.8370723575196078} 12.08 & \cellcolor[rgb]{0.9781854635254902,0.8875721630666666,0.8432079741549019} 0.13 \\
\bottomrule
\end{tabular}
}\caption{\label{tab:mistral_gen_results} PRR$\uparrow$ 50\% with various generation quality metrics for UQ methods in selective generation tasks with the \Mistral v0.2 model. Warmer color indicates better results.}\end{table*}

 \newpage
 \clearpage
 
 \section{Additional Experimental Results with Uncertainty Normalization}
 \label{sec:normalization_appendix}

   \begin{table}[h] 
\scriptsize
%\footnotesize
\centering
\resizebox{0.6\textwidth}{!}{
\begin{tabular}{l|r|r|r|r}
\toprule
{\textbf{UQ Method}} &  \textbf{Linear} &  \textbf{Quantile} &  \textbf{Binned PCC} &  \textbf{Isotonic PCC} \\
\midrule
Maximum Sequence Probability & 0.000 & 0.000 & -0.041 & 0.000 \\
Perplexity & 0.000 & 0.000 & 0.016 & 0.001 \\
Mean Token Entropy & 0.000 & 0.000 & 0.002 & 0.000 \\
% MonteCarloSequenceEntropy & 0.000 & 0.000 & -0.045 & 0.000 \\
% MonteCarloNormalizedSequenceEntropy & 0.000 & 0.000 & 0.005 & 0.000 \\
Mean Pointwise Mutual Information & 0.000 & 0.000 & -0.398 & 0.000 \\
Mean Conditional PMI & 0.000 & 0.000 & -0.298 & 0.000 \\
Rényi Divergence & 0.000 & 0.000 & -0.121 & 0.000 \\
Fisher-Rao Distance& 0.000 & 0.000 & -0.125 & 0.000 \\
TokenSAR & 0.000 & 0.000 & 0.022 & 0.000 \\
CCP & 0.000 & 0.000 & -0.002 & -0.001 \\
Monte Carlo Sequence Entropy & 0.000 & 0.000 & -0.045 & 0.000 \\
Monte Carlo Normalized Sequence Entropy & 0.000 & 0.000 & 0.005 & 0.000 \\
Semantic Entropy & 0.000 & 0.000 & -0.023 & 0.000 \\
Sentence SAR & 0.000 & 0.000 & -0.008 & 0.000 \\
SAR & 0.000 & 0.000 & 0.001 & 0.000 \\
\PT & 0.000 & 0.000 & -0.784 & -0.001 \\
NumSet & 0.000 & 0.000 & 0.375 & 0.000 \\
EigValLaplacian NLI Score Entail & 0.000 & 0.000 & 0.009 & 0.001 \\
EigValLaplacian Jaccard Score & 0.000 & -0.002 & 0.000 & 0.000 \\
DegMat NLI Score Entail & 0.000 & 0.000 & 0.011 & 0.001 \\
DegMat Jaccard Score & 0.000 & 0.001 & 0.060 & 0.000 \\
Eccentricity NLI Score Entail & 0.000 & 0.000 & 0.023 & 0.000 \\
Eccentricity Jaccard Score & 0.001 & 0.002 & 0.092 & 0.031 \\
Lexical Similarity Rouge-L & 0.000 & 0.000 & 0.073 & 0.000 \\
\bottomrule
\end{tabular}
}
\caption{\label{tab:total_confidence_prr} The difference between PRR of raw uncertainty and bounded confidence obtained with various normalization techniques. The lower is better; negative values represent cases when confidence performs better than raw uncertainty scores.}\end{table}

 \section{LLM Text Generation Quality}
 \label{sec:quality_metrics}
    
   \begin{table*}[!h] 
\centering
\resizebox{0.95\textwidth}{!}{
\begin{tabular}{c|c|c|c|c|cc|cc|cc}

\toprule
\multirow{2}{*}{\textbf{Model}} & \multicolumn{1}{c|}{\textbf{CoQA}} & \multicolumn{1}{c|}{\textbf{TriviaQA}} & \multicolumn{1}{c|}{\textbf{MMLU}} & \multicolumn{1}{c|}{\textbf{GSM8k}} & \multicolumn{2}{c|}{\textbf{XSum}} & \multicolumn{2}{c|}{\textbf{WMT14 Fr-En}} & \multicolumn{2}{c}{\textbf{WMT19 De-En}} \\ \cline{2-11}
    & \textbf{AlignScore} & \textbf{AlignScore} &\textbf{Accuracy} & \textbf{Accuracy} & \textbf{Rouge-L} & \textbf{AlignScore} &\textbf{Comet} & \textbf{AlignScore} &\textbf{Comet} & \textbf{AlignScore} \\\midrule
\Stable v2 & 0.77 & 0.69  & 0.57 & 0.55 & 0.21 & 0.03 & 0.87 & 0.85 & 0.88 & 0.84 \\
\Mistral v0.2 & 0.79 & 0.74 & 0.64 & 0.38 & 0.23 & 0.07 & 0.86 & 0.84 & 0.86 & 0.81 \\
\bottomrule
\end{tabular}
}\caption{\label{tab:stable_mistral} Generation quality metrics for LLMs without instruction tuning.}\end{table*}

\begin{table*}[!ht]
\centering
\normalsize
\resizebox{0.7\textwidth}{!}{
\begin{tabular}{l|l|c|c|c}
\toprule
 \multirow{2}{*}{\textbf{Model}} & \multirow{2}{*}{\textbf{Prompt}} &  \multicolumn{1}{c|}{\textbf{CoQA}} & \multicolumn{1}{c|}{\textbf{TriviaQA}} & \multicolumn{1}{c}{\textbf{MMLU}} \\ \cline{3-5}
& & \textbf{AlignScore} & \textbf{AlignScore} & \textbf{Accuracy} \\\midrule
\multirow{7}{*}{\textbf{GPT-4o-mini}} & Linguistic1S & 0.74 & 0.81 & 0.75 \\
& Verbalized 1S top-1 & 0.73 & 0.81 & 0.75 \\
& Verbalized 1S top-k & 0.70 & 0.80 & 0.75 \\
& Verbalized 2S CoT & 0.71 & 0.84 & 0.79 \\
& Verbalized 2S top-1 & 0.74 & 0.81 & 0.76 \\
& Verbalized 2S top-k & 0.69 & 0.80 & 0.76 \\
& Base & 0.74 & 0.81 & 0.76 \\
\midrule
\multirow{7}{*}{\textbf{\Stable Chat}} &  Linguistic1S & 0.68 & 0.74 & 0.49 \\
& Verbalized 1S top-1 & 0.69 & 0.73 & 0.52 \\
& Verbalized 1S top-k & 0.56 & 0.74 & 0.55 \\
& Verbalized 2S CoT & 0.66 & 0.74 & 0.43 \\
& Verbalized 2S top-1 & 0.69 & 0.73 & 0.51 \\
& Verbalized 2S top-k & 0.69 & 0.74 & 0.56 \\
& Base & 0.69 & 0.73 & 0.51 \\
\midrule
\multirow{7}{*}{\textbf{\Mistral v0.2 Instruct}} & Linguistic1S & 0.69 & 0.71 & 0.58 \\
& Verbalized 1S top-1 & 0.69 & 0.69 & 0.58 \\
& Verbalized 1S top-k & 0.67 & 0.70 & 0.56 \\
& Verbalized 2S CoT &  0.60 & 0.69 & 0.54 \\
& Verbalized 2S top-1 & 0.70 & 0.70 & 0.57 \\
& Verbalized 2S top-k & 0.68 & 0.70 & 0.56 \\
& Base & 0.70 & 0.70 & 0.57 \\
\bottomrule
\end{tabular}
}
\caption{\label{tab:blackbox_quality}Generation quality for instruction-tuned LLMs.}
\end{table*}

 \end{document}